\documentclass[preprint,12pt]{elsarticle}
\usepackage[pdfborder={0 0 0}]{hyperref}
\usepackage{amsthm,amsmath,amsfonts,latexsym,amssymb}
\usepackage{graphics,fancyhdr,graphicx}
\usepackage[utf8]{inputenc}
\usepackage[margin=1in]{geometry}
\usepackage{multirow,booktabs}
\usepackage[table]{xcolor}
\usepackage{listings}
\usepackage{tabularx}
\usepackage{placeins}
\usepackage{comment}
\usepackage{lineno}
\usepackage{lscape}
\usepackage{bm}
\usepackage{hyperref}
\usepackage{mathrsfs}
\usepackage{arydshln}
\usepackage{appendix}
\definecolor{custom-blue}{RGB}{3,69,173}
\hypersetup{
    colorlinks = true,
    urlcolor   = blue,
    citecolor  = custom-blue,
}
\biboptions{numbers,sort&compress}
\definecolor{listinggray}{gray}{0.9}
\definecolor{lbcolor}{rgb}{0.9,0.9,0.9}
\definecolor{Darkgreen}{RGB}{0,100,0}
\usepackage{caption}

\usepackage{float}
\usepackage{booktabs}
\usepackage{multirow}
\usepackage{tabularray}
\usepackage{algpseudocode}
\usepackage{mathtools}
\usepackage{subcaption}
\usepackage{enumitem}
\usepackage{algpseudocode}
\usepackage{algorithm}
\usepackage{rotating}
\usepackage{booktabs}
\usepackage{array}
\usepackage{hyperref}
\usepackage{amsmath}
\usepackage{graphicx}
\usepackage{booktabs}
\usepackage{array}
\usepackage{comment}

\newcommand{\bx}{\boldsymbol{x}}
\usepackage{bm}
\newtheoremstyle{remarkstyle}  % Name
  {5pt}                        % Space above
  {5pt}                        % Space below
  {}                           % Body font (upright)
  {}                           % Indent amount
  {\bfseries}                  % Theorem head font (bold)
  {.}                          % Punctuation after theorem head
  { }                          % Space after theorem head
  {}                           % Theorem head spec (empty = `normal`)

\begin{document}

\makeatletter
\def\ps@pprintTitle{%
  \let\@oddhead\@empty
  \let\@evenhead\@empty
  \let\@oddfoot\@empty
  \let\@evenfoot\@oddfoot
}
\makeatother

\abovedisplayskip=6.0pt
\belowdisplayskip=6.0pt

\begin{frontmatter}

\title{Efficient Training of Deep Neural Operator Networks via Randomized Sampling}

\author[1]{Sharmila Karumuri}
\author[1]{Lori Graham-Brady}
\author[1]{Somdatta Goswami\corref{cor1}}
\ead{sgoswam4@jhu.edu}

\address[1]{Department of Civil and Systems Engineering, Johns Hopkins University, U.S.A.}

\cortext[cor1]{Corresponding author.}

\begin{abstract}
\noindent
Neural operators (NOs) employ deep neural networks to learn mappings between infinite-dimensional function spaces. Deep operator network (DeepONet), a popular NO architecture, has demonstrated success in the real-time prediction of complex dynamics across various scientific and engineering applications. 
In this work, we introduce a random sampling technique to be adopted during the training of DeepONet, aimed at improving the generalization ability of the model, while significantly reducing the computational time. 
The proposed approach targets the trunk network of the DeepONet model that outputs the basis functions corresponding to the spatiotemporal locations of the bounded domain on which the physical system is defined. 
While constructing the loss function, DeepONet training traditionally considers a uniform grid of spatiotemporal points at which all the output functions are evaluated for each iteration. 
This approach leads to a larger batch size, resulting in poor generalization and increased memory demands, due to the limitations of the stochastic gradient descent (SGD) optimizer. 
The proposed random sampling over the inputs of the trunk net mitigates these challenges, improving generalization and reducing memory requirements during training, resulting in significant computational gains. 
We validate our hypothesis through three benchmark examples, demonstrating substantial reductions in training time while achieving comparable or lower overall test errors relative to the traditional training approach. Our results indicate that incorporating randomization in the trunk network inputs during training enhances the efficiency and robustness of DeepONet, offering a promising avenue for improving the framework's performance in modeling complex physical systems.
\end{abstract}
\end{frontmatter}

\section{Introduction}\label{sec1.2}
\noindent

Real-time prediction of complex phenomena is critical across various scientific and engineering disciplines. The ability to rapidly and accurately forecast the behavior of complex systems is essential for informed decision-making, risk assessment, and optimization in various fields.
Traditionally, the modeling and simulation of these complex phenomena have relied heavily on numerical solvers for partial differential equations (PDEs), such as the Finite Element Method (FEM) or Finite Difference Method (FDM), which are computationally intensive and involve time-consuming workflows. Recent advancements in machine learning, particularly in deep learning frameworks, have revolutionized predictions for complex physical systems. These developments have paved the way for data-driven surrogate models, offering near-instantaneous predictions and the ability to generalize across similar scenarios. 

A significant milestone in the domain of modeling physical systems with deep learning techniques was the development of physics-informed neural networks (PINNs) \cite{raissi2019physics}. Despite their effectiveness, PINNs are trained for specific boundary and initial conditions, as well as loading or source terms, and require expensive training. 
Therefore, they are not particularly effective for other operating conditions and real-time inference, although transfer learning can somewhat alleviate this limitation. Alternatives such as learning function-to-function mapping using residual neural networks and the governing physics have been proposed \cite{karumuri2020simulator}. However, there remains a need for a more generalized architecture that could handle complex engineering scenarios for many different boundary/initial conditions and loading conditions, without further training or perhaps with very light training.

\begin{figure}[H]
\centering
\includegraphics[width=0.8\textwidth]{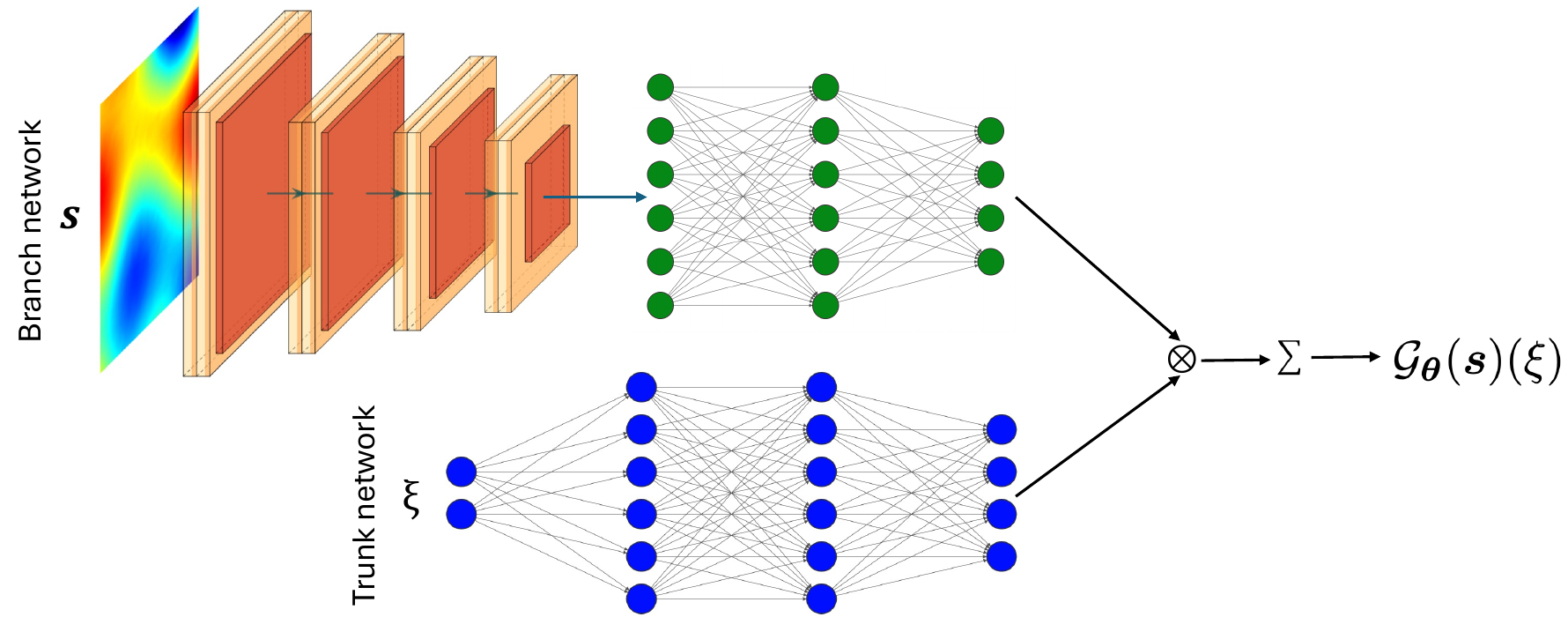}
\caption{A schematic representation of the DeepONet architecture. The branch network takes as input the functional field $\boldsymbol{s}$, while the trunk network takes as input the coordinates at which the output field is to be evaluated.}
\label{fig:DON_schematic}
\end{figure}

The introduction of neural operators (NOs), particularly the deep operator networks (DeepONet) proposed by Lu et al. in 2021 \cite{lu2021learning}, effectively met this need. DeepONet has been formulated based on the universal approximation theorem of Chen \& Chen \cite{chen1995universal}, allowing the mapping between infinite dimensional functions using deep neural networks (DNNs).
The architecture of DeepONet comprises two sub-networks: the branch and the trunk, as illustrated in Figure \ref{fig:DON_schematic}. 
The branch network is designed to take as inputs the varying conditions of the PDE evaluated at fixed sensor locations and learn the coefficients. The trunk network takes the spatial and temporal evaluation locations and learns the basis functions. The solution operator is written as the inner product of the outputs of the branch and the trunk networks.

\begin{figure}[H]
\centering
\includegraphics[width=4.0in]{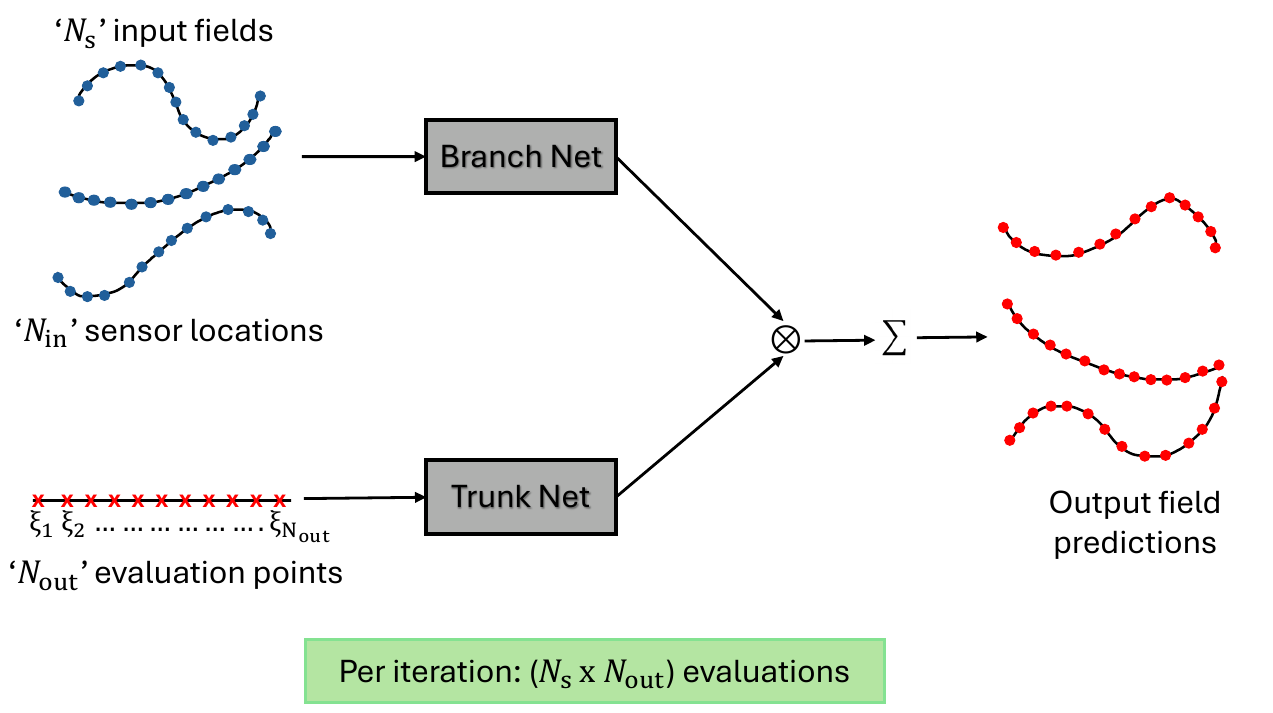}\\
(a)\\[1em] % Adds space between the images
\includegraphics[width=4.0in]{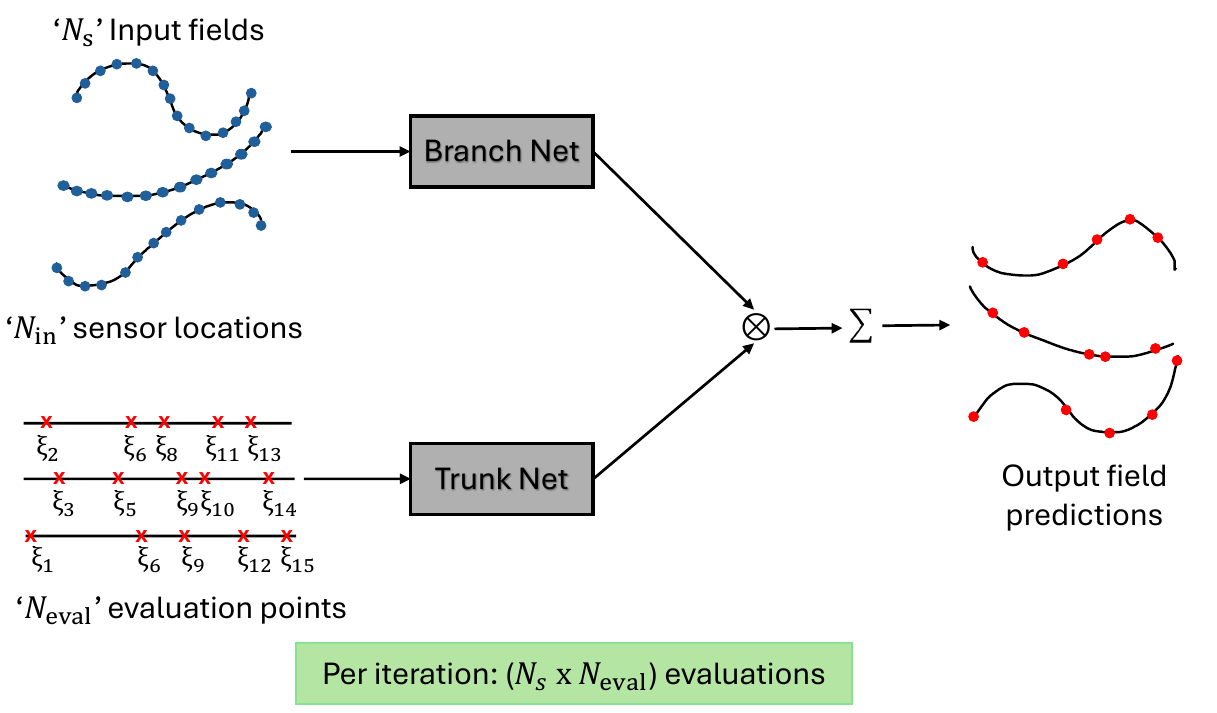}\\
(b)
\caption{Training approaches for DeepONet: (a) traditional, and (b) randomized (ours), for $N_s = 3$ input fields.}
\label{fig:Approaches}
\end{figure}

Traditionally, the training of DeepONet involves the evaluation of the trunk network for all spatio-temporal locations at which the ground truth is available for every input function considered in the branch network, to construct the loss function (as illustrated in Figure~\ref{fig:Approaches}(a)). However, this approach results in larger effective batch sizes, leading to several challenges: poor generalization, slow convergence, and large memory requirements. These issues arise from the limitations of the stochastic gradient descent (SGD) optimizer, which updates the network's parameters (weights and biases) during backpropagation. Moreover, the training process is highly sensitive to the learning rate, further complicating efforts to achieve optimal performance.

Unfortunately, these large batch sizes during DeepONet training often exacerbate these challenges, leading to reduced test set accuracy, as observed in several studies \cite{lecun2002efficient, keskar2016large, goyal2017accurate} carried out on DNNs. \cite{keskar2016large} conducted a detailed investigation into the effects of batch size on the generalization performance of deep learning models. The authors highlighted that larger batch sizes tend to converge to sharp minima of the training and testing functions, which can lead to poorer generalization. In contrast, smaller batch sizes consistently converge to flatter minima, resulting in better generalization performance. They also noted that the differences in generalization performance due to batch size were consistent across a range of network architectures and problem domains. Furthermore, they hypothesize that the larger estimation noise present in small-batch training encourages the weights to escape from sharp minima and move towards flatter minima, which are believed to generalize better. These insights underscore the importance of carefully considering batch size when training DeepONet models, as it can significantly impact the model's ability to generalize and its overall performance on unseen data.

To address the issues related to the reduced generalization ability of DeepONet when trained with large batch sizes, we propose a random sampling technique for the trunk network. Our proposed method is particularly advantageous in scenarios involving mini-batch training on large real-world datasets, where accuracy is highly sensitive to batch size. Specifically, in each training iteration, for a given functional input to the branch network, the solution operator is evaluated in randomly selected locations rather than at all locations where the ground truth is available within the interior domain (see Figure~\ref{fig:Approaches}(b)). 
The randomization of trunk net inputs leads to decreased generalization errors due to the reduced effective batch size, as well as a decreased training time. 
While similar randomization approaches have been explored in recent literature \cite{di2023neural, hao2023instability, borrel2024sound}, our work distinguishes itself by conducting a comprehensive analysis of how randomizing the trunk network inputs affects generalization capabilities, convergence rates, and overall training efficiency in DeepONet models. Furthermore, our proposed random sampling technique is versatile and applicable across various modified DeepONet frameworks \cite{bahmani2024resolution,cao2024deep,kumar2024synergistic,he2023novel,jin2022mionet,haghighat2024deeponet,taccari2024developing,he2024sequential} including the physics-informed frameworks \cite{mandl2024separable,goswami2023physics,goswami2022physics,wang2021learning}, designed to address problem-specific challenges. Of all the modified architectures of DeepONet, this approach does not apply to POD-DeepONet \cite{kontolati2023influence} and two-stage training of the DeepONet framework \cite{lee2024training} due to architectural restrictions. 

The chapter is organized as follows: In Sec.~\ref{sec1.3}, we introduce the proposed random sampling scheme for training DeepONet and provide a detailed algorithm for implementing this scheme. 
In Sec.~\ref{sec1.4}, we present a comprehensive evaluation of our sampling methodology, applying it to three benchmark examples widely recognized in the literature. For each case study, we conduct ablation experiments by systematically varying two critical parameters: (1) the number of training data points and (2) the number of evaluation points utilized in the trunk network during each iteration. Through this methodical analysis, we evaluate the performance of our proposed framework in terms of both generalization error and computational efficiency. Finally, we summarize our observations and provide concluding remarks in Sec.~\ref{sec1.5}.

\section{Methodology}\label{sec1.3}
\noindent

In data-driven operator learning, the objective is to learn a mapping between two Banach spaces of vector-valued functions, $\mathcal{S}$ and $\mathcal{U}$, defined as follows:
\begin{equation}
\begin{split} \mathcal{S} = \{ s : \mathcal{I} \rightarrow \mathbb{R}^{d_s} \}, \quad \mathcal{I} \subseteq \mathbb{R}^{d_i},\\
\mathcal{U} = \{ u : \mathcal{O} \rightarrow \mathbb{R}^{d_u} \}, \quad \mathcal{O} \subseteq \mathbb{R}^{d_o}, 
\end{split}
\end{equation}
here, $\mathcal{S}$ represents the space of input functions, and $\mathcal{U}$ denotes the space of corresponding output functions.
The task of operator learning is to approximate the nonlinear operator $\mathcal{G}: \mathcal{S} \rightarrow \mathcal{U}$, i.e., $u = \mathcal{G}(s)$, through a parametric mapping:
\begin{equation}
\mathcal{G} : \mathcal{S} \times \bm{\Theta} \rightarrow \mathcal{U} \quad \text{or} \quad \mathcal{G}_{\bm{\theta}} : \mathcal{S} \rightarrow \mathcal{U}, \quad \bm{\theta} \in \bm{\Theta}
\end{equation}
where $\bm{\Theta}$ is a finite-dimensional parameter space.
The optimal parameters can be estimated as follows:
\begin{equation}
\arg\min_{\bm{\theta}} \frac{1}{N_\text{train}} \sum_{i=1}^{N_\text{train}} \left| u_i - \mathcal{G}_{\bm{\theta}}(s_i) \right|^2
\end{equation}
where the dataset $\mathcal{D}_{OL} = \{(s_i, u_i)\}_{i=1}^{N_\text{train}}$ contains $N_\text{train}$ pairs of input-output functions.
Specifically, when dealing with physical systems described by PDEs one of the functions, such as the variable coefficients, forcing term, source term, initial condition, and boundary conditions, constitute the input functions in the space $\mathcal{S}$. 
The corresponding output functions represent the solution of the PDE, which resides in the space $\mathcal{U}$, and the goal is to learn the mapping from different inputs to these solutions.

In practice, we typically only have access to a finite set of observations of these input-output functions, corresponding to their discretized versions. 
In the general sense, this means that, the $i$-th input function $s_i(\eta)$ is discretized at $N_{\text{in}}$ input sensor locations and stored as a vector $\bm{s}_i = [s_i(\eta_1), s_i(\eta_2), \dots, s_i(\eta_{N_{\text{in}}})]$, where $\eta \in \mathcal{I}$ and $\bm{s} \in \mathcal{S}$.
Similarly, the corresponding output function $u_i(\xi)$ is discretized at $N_{\text{out}}$  output sensor locations and stored as a vector $\bm{u}_i = [u_i(\xi_1), u_i(\xi_2), \dots, u_i(\xi_{N_{\text{out}}})]$, where $\xi \in \mathcal{O}$ and $\bm{u} \in \mathcal{U}$.
This discretized general data configuration $\mathcal{D}_\text{train} = \{(\bm s_i,\bm u_i) \}_{i=1}^{N_\text{train}}$ is then used for operator learning, often employing models like DeepONet.

DeepONet is based on the universal approximation theorem for operators \cite{chen1995universal}.
The standard DeepONet architecture consists of two DNNs: the branch and trunk networks. 
The branch network takes as input the input functions in $\mathcal{S}$, discretized at $N_{\text{in}}$ input sensor locations $[\eta_1, \eta_2, \dots, \eta_{N_{\text{in}}}]$ to produce $p$ coefficients. 
Meanwhile, the trunk network
takes in the output senor locations $\xi$, to generate corresponding basis functions. 
The parameterized functional $\mathcal{G}_{\bm{\theta}}(s)(\xi)$, where $\mathcal{G}(s)(\xi) \approx \mathcal{G}_{\bm{\theta}}(s)(\xi)$ is represented using DeepONet as follows:
\begin{equation}
\label{eq:output_deeponets}
    \begin{split} \mathcal{G}_{\bm{\theta}}(s)(\xi) \approx \mathcal{G}_{\bm{\theta}}(\bm s)(\xi)&{=} \sum_{k=1}^p br_k \cdot tr_k \\
      &{=} \sum_{k=1}^{p} br_k([s(\eta_1), s(\eta_2), \dots,s(\eta_{N_{\text{in}}})]; \bm{\theta}) \cdot tr_k(\xi; \bm{\theta}),   
\end{split}
\end{equation}
where $\{br_1, br_2, \dots, br_p\}$ are the outputs of the branch network, and $\{tr_1, tr_2, \dots, tr_p\}$ are the outputs of the trunk network and $\bm{\theta}$ encompasses the trainable parameters of both the branch and trunk networks.
The schematic representation of DeepONet is shown in  Figure~\ref{fig:DON_schematic}.

The trainable parameters, $\bm{\theta}$, of this DeepONet model, in a data-driven setting, are traditionally learned by minimizing a loss function expressed as:
\begin{equation}
\label{eq:loss_func_deeponet}
\mathcal{L}(\bm{\theta}) = \frac{1}{N_s N_\text{out}}\sum_{i = 1}^{N_s} \sum_{j = 1}^{N_\text{out}} \left(u_i(\xi_j) - \mathcal{G}_{\bm{\theta}}(\bm s_i)(\xi_j)\right)^2,
\end{equation}
where $N_s$ refers to the number of input functions in a given batch (i.e., the batch size, denoted as $bs$), and $N_\text{out}$ represents the number of output sensor locations where the output function data is available for each input function.

In Eq.~\ref{eq:loss_func_deeponet}, $u_i(\xi_j)$ refers to the ground truth of the output function and $\mathcal{G}_{\bm{\theta}}(\bm s_i)(\xi_j)$ is the DeepONet prediction. The weights are updated using any stochastic gradient descent method, such that,
\begin{equation}
    \bm{\theta}_{r+1} = \bm{\theta}_{r} - \alpha\nabla_{\bm{\theta}}\mathcal L(\bm{\theta}_r),
\end{equation}
where $\alpha$ is the learning rate and $\nabla$ denotes the gradient.

However, this approach of evaluating each output function in a batch across all points in the domain, as detailed in Eq.~\eqref{eq:loss_func_deeponet}, leads to larger effective batch sizes during training, specifically $N_s \times N_\text{out}$ for each iteration (see Figure~\ref{fig:Approaches}(a)).
These large batch sizes often lead to poor generalization errors, slower convergence, and increased memory requirements. 

\begin{figure}[H]
    \centering
    \includegraphics[width=0.8\textwidth]{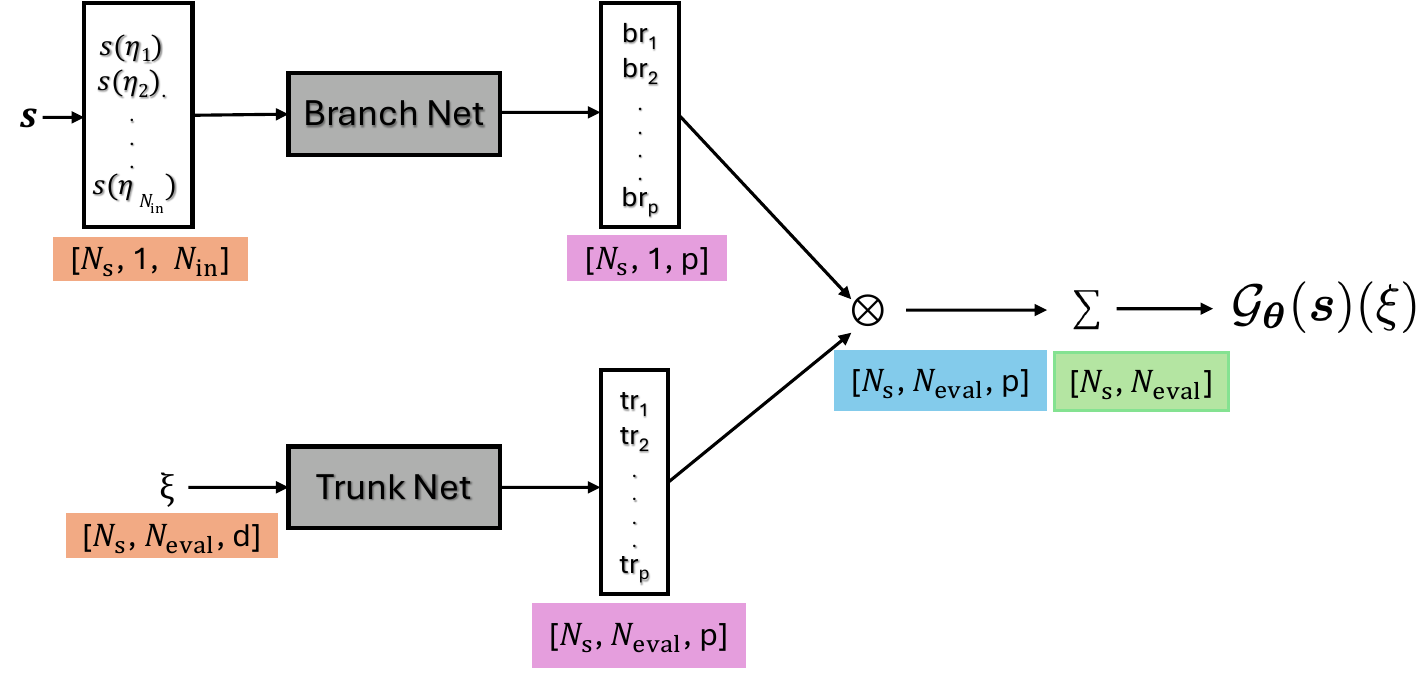}
    \caption{Our DeepONet Architecture: This figure illustrates our DeepONet architecture, which employs batch-based forward passes. In each iteration, $N_s$ input functions sampled at $N_{\text{in}}$ sensors and $N_{\text{eval}}$ input coordinates of $\xi$ (which are $d$-dimensional) are randomly sampled as inputs to the trunk network. Both networks, consistent with traditional DeepONet, produce a latent dimension of $p$. The key operations of the architecture include the outer product combining branch and trunk batches, and the final summation over the latent dimension $p$, resulting in $[N_s, N_{\text{eval}}]$ output function evaluations per iteration.}
    \label{fig:Architecture}
\end{figure}

In this work, we propose a strategy to alleviate the challenges of poor generalization caused by larger batch sizes, by adopting a random sampling scheme for evaluation points that serve as inputs to the trunk network.
Specifically, in each training iteration, for a given functional input to the branch network, the output is evaluated at randomly selected points ($N_\text{eval}$), rather than at all points at which the ground truth is available (see Figure~\ref{fig:Approaches}(b)).
\begin{equation}
\label{eq:loss_func_randdeeponet}
\mathcal{L}(\bm{\theta}) = \frac{1}{N_s N_\text{eval}}\sum_{i = 1}^{N_s} \sum_{j = 1}^{N_\text{eval}} \left(u_i(\xi_{ij}^\ast) - \mathcal{G}_{\bm{\theta}}(\bm s_i)(\xi_{ij}^\ast)\right)^2,
\end{equation}
where $\xi_{ij}^\ast$ are randomly selected evaluation points and $N_\text{eval}$ is the number of evaluation points.
This randomization of the inputs to the trunk network leads to decreased generalization errors, due to the reduced effective batch size $(= N_s \times N_\text{eval})$, which lowers the training time. 
The training process for DeepONet using our randomized sampling approach is outlined in detail in Algorithm~\ref{alg:DeepONet_training} and the network structure with batch tracing is shown in Figure~\ref{fig:Architecture}.
Note that \cite{lu2022comprehensive} introduced an efficient training approach that evaluates the trunk network only once for all $N_\text{out}$ sensors. 
This method employs the Einstein summation operation (Einsum) to compute the dot product, resulting in $N_s \times N_\text{out}$ output evaluations.
Although this approach significantly reduces computational costs, challenges related to managing larger effective batch sizes remain substantial.

\begin{algorithm}
\caption{The proposed sampling technique to train DeepONet.}  
\label{alg:DeepONet_training}
\begin{algorithmic}[1]
\Require Branch and trunk networks architecture, number of training samples $N_\text{train}$, training data $\mathcal{D}_\text{train} = \{(\bm s_i,\bm u_i) \}_{i=1}^{N_\text{train}}$, where
$\bm s_i = [ s_i(\eta_1), s_i(\eta_2), \dots, s_i(\eta_{N_{\text{in}}}) ]$ and 
$\bm u_i = [u_i(\xi_1), u_i(\xi_2), \dots, u_i(\xi_{N_\text{out}})]$,
output sensor locations $\Xi = [\xi_1, \xi_2, \dots, \xi_{N_\text{out}}]$,
batch size $\text{bs}$,
number of batches $N_\text{batch} = \lceil \frac{N_\text{train}}{\text{bs}} \rceil$,
number of evaluation points $N_\text{eval}$,
learning rate $\alpha$, and number of epochs $N_\text{epochs}$.
\State Initialize parameters of branch and trunk networks $\bm{\theta}$.
\For{$k = 1$ \textbf{to} $N_\text{epochs}$}
    \State \textbf{Shuffle training data}: $\mathcal{D}_\text{train} \gets \{ (\bm s_{\sigma(i)}, \bm u_{\sigma(i)}) \}_{i=1}^{N_\text{train}}$, where $\sigma$ is a permutation.
    \For{$j = 1$ \textbf{to} $N_\text{batch}$}
        \State \text{start }$ \gets (j-1)\times \text{bs} + 1$, \text{end }$ \gets \min(j \times\text{bs}, N_\text{train})$
        \State \textbf{Get mini-batch}: $\mathcal{D}_{\text{train}, j} \gets \{(\bm s_a,\bm u_a) \}_{a=\text{start}}^{\text{end}}$ 
        \For{$a = \text{start}$ \textbf{to} $\text{end}$}
            \State Get  $\Xi^\ast_a \subset \Xi$ such that $|\Xi^\ast_a| = N_\text{eval}$.
            \State Select $N_\text{eval}$ distinct indices $\{i_{(a, 1)}, i_{(a, 2)}, \ldots, i_{(a, N_\text{eval})}\}$ uniformly at random from $\{1, 2, \ldots, N_{\text{out}}\}$ with $i_{(a, p)} \neq i_{(a, q)}$ for $p \neq q$.
            \State $\Xi^\ast_a = [ \xi_{i_{(a, 1)}}, \xi_{i_{(a, 2)}}, \ldots, \xi_{i_{(a, N_\text{eval})}} ]$
        \EndFor
        \State Compute loss:
        $$
        \mathcal{L}(\bm{\theta}) = \frac{1}{(\text{end} - \text{start} + 1) \cdot N_\text{eval}} \sum_{a = \text{start}}^{\text{end}} \sum_{l=1}^{N_\text{eval}} \Bigg( u_a(\xi_{i_{(a, l)}}) - \mathcal{G}_{\bm{\theta}}(\bm{s}_a)(\xi_{i_{(a, l)}}) \Bigg)^2
        $$
        \State Update parameters by computing $\nabla_{\bm{\theta}} \mathcal{L}(\bm{\theta})$ (backpropagation)
    \EndFor
\EndFor
\State \Return $\bm{\theta}^{\ast}$ \Comment{Return trained DeepONet parameters.}
\end{algorithmic}
\end{algorithm}

\section{Results}\label{sec1.4}
\noindent

In this section, we demonstrate the effectiveness of the proposed randomized sampling approach for training DeepONet on benchmark numerical examples taken from the literature.
A summary of the examples considered in this work is presented in Table~\ref{tab:examples}. 
The architecture of DeepONet as well as the hyperparameters used to generate the results presented in this work are reported in Table~\ref{tab:architectures-hyperparameters}. 
The code and data used for all the examples are made available at \url{https://github.com/Centrum-IntelliPhysics/Efficient_DeepONet_training}.
\begin{table}[H]
\centering
\scriptsize
\caption{A schematic representation of the operator learning benchmarks under consideration in this work.}
\begin{tabular}{p{1.5cm} p{4.5cm} p{4.5cm} p{4.5cm}}
\toprule
\textbf{Case} & \textbf{Dynamical System} & \textbf{Diffusion-reaction} & \textbf{Heat Equation} \\ 
\midrule
\textbf{Model Output} & 
$\begin{aligned}
&\frac{du}{dt} = s(t),\\
&u(0) = 0\text{ and } t\in[0,1]\\
&\mathcal{G}_{\boldsymbol{\theta}}: s(t) \to u(t).
\end{aligned}$ &
$\begin{aligned}
&\frac{\partial u}{\partial t} = D \frac{\partial^2 u}{\partial x^2} + k u^2 + s(x), \\
&D=0.01, \ k=0.01, \\
&(t, x) \in (0, 1] \times (0, 1],\\
&u(0, x) = 0, \ x \in (0,1)\\
&u(t, 0) = 0, \ t \in (0,1)\\
&u(t, 1) = 0, \ t \in (0,1)\\
&\mathcal G_{\boldsymbol{\theta}}: s(x) \to u(t,  x).
\end{aligned}$ &
$\begin{aligned}
&-\nabla \cdot (a(\bx) \nabla u(\bx)) = 0,  \\
&\bx = (x_1, x_2), \\
&\bx  \in \Omega = [0, 1]^2,\\
&u(0,x_2) = 1,\ u(1,x_2) = 0, \\ 
&\frac{\partial u (x_1,0)}{\partial n}= \frac{\partial u(x_1, 1)}{\partial n}= 0, \\ 
&\mathcal G_{\boldsymbol{\theta}}: a(\bx) \to u(\bx).
\end{aligned}$ \\ 
\midrule
\textbf{Input Function} & 
$\begin{aligned}
&s(t) \sim \mathrm{GP}(0, k(t,t')), \\
&\ell_t = 0.2, \ \sigma^2 = 1.0 , \ t\in[0,1]\\
&k(t, t') = \sigma^2 \exp\left\{- \frac{\|t - t'\|^2}{2\ell_t^2}\right\}.
\end{aligned}$ &
$\begin{aligned}
&s(x) \sim \mathrm{GP}(0, k(x, x')), \\
&\ell_{x} = 0.2, \ \sigma^2 = 1.0,\\
&k(x, x') = \sigma^2 \exp\left\{- \frac{\|x - x'\|^2}{2\ell_x^2}\right\}.
\end{aligned}$ &
$\begin{aligned}
&\log (a(\bx)) \sim \mathrm{GP}(\mu(\bx), k(\bx, \bx'))\\
&\mu(\bx)= 0, \ \ell_{x_1} = 0.1, \ \ell_{x_2} = 0.15, \ \sigma^2 = 1.0\\
&k(\bx, \bx') = \sigma^2 \exp\left\{- \sum_{i=1}^{2} \frac{\|x_{i} - x'_{i}\|^2}{2\ell_{x_i}^2}\right\}.
\end{aligned}$ \\ 
\midrule
\textbf{Samples} & 
\begin{minipage}{4.5cm}
    \centering
    \hspace{-0.8cm}
    \includegraphics[width=0.6\linewidth]{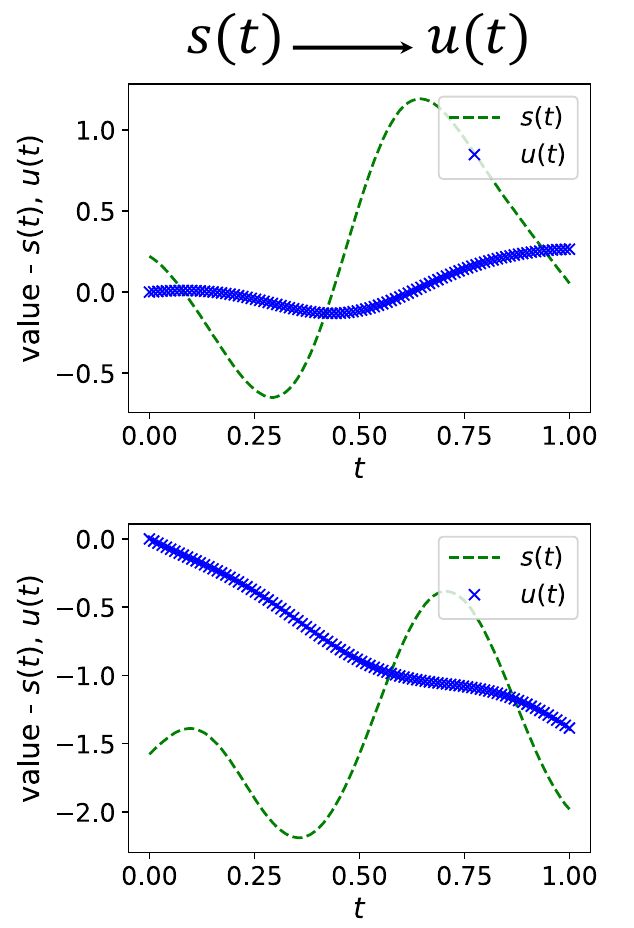}
\end{minipage} &
\begin{minipage}{4.5cm}
    \centering
    \hspace{-0.8cm}
    \includegraphics[width=1.1\linewidth]{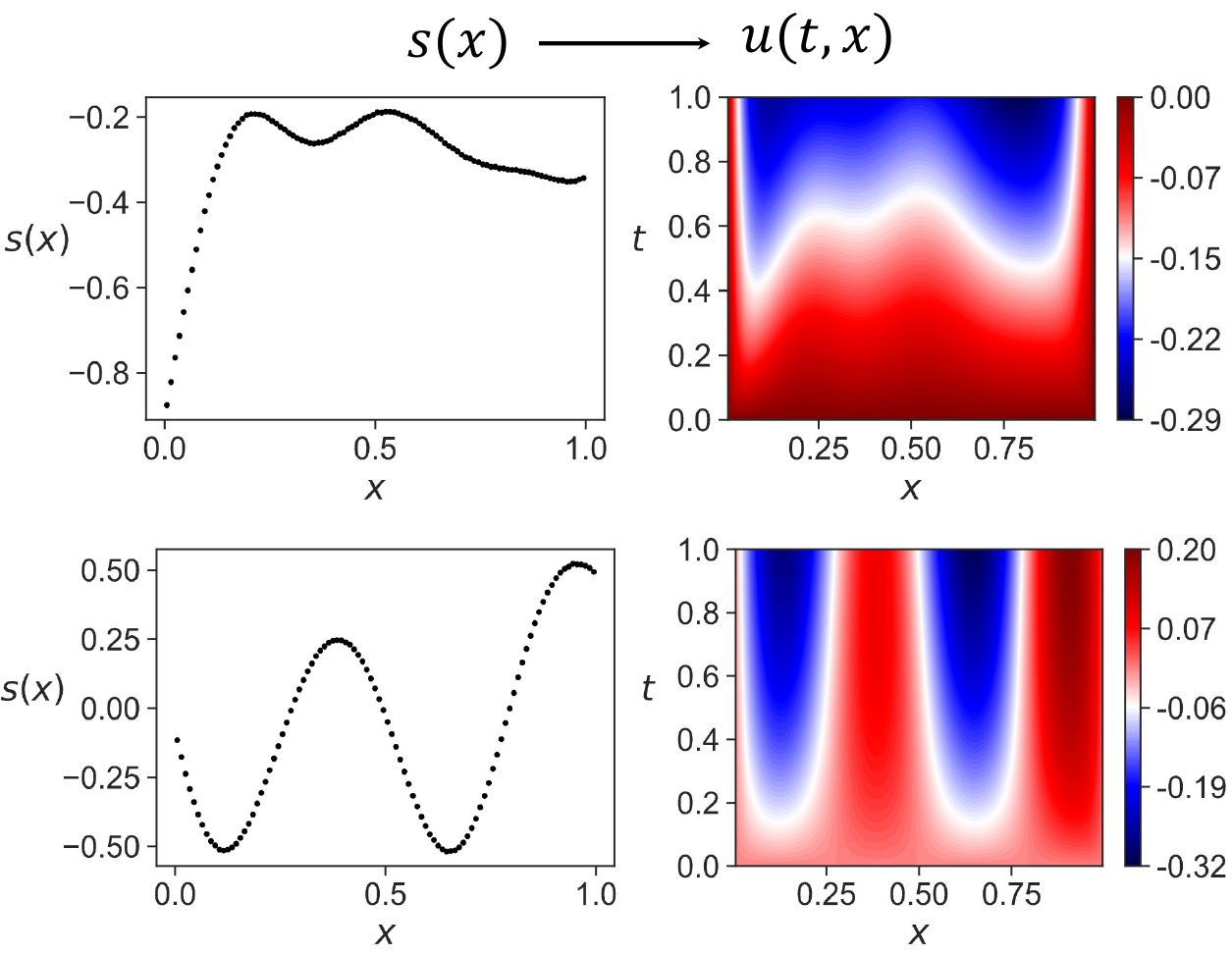} 
\end{minipage} &
\begin{minipage}{4.5cm}
    \centering
    \includegraphics[width=1.1\linewidth]{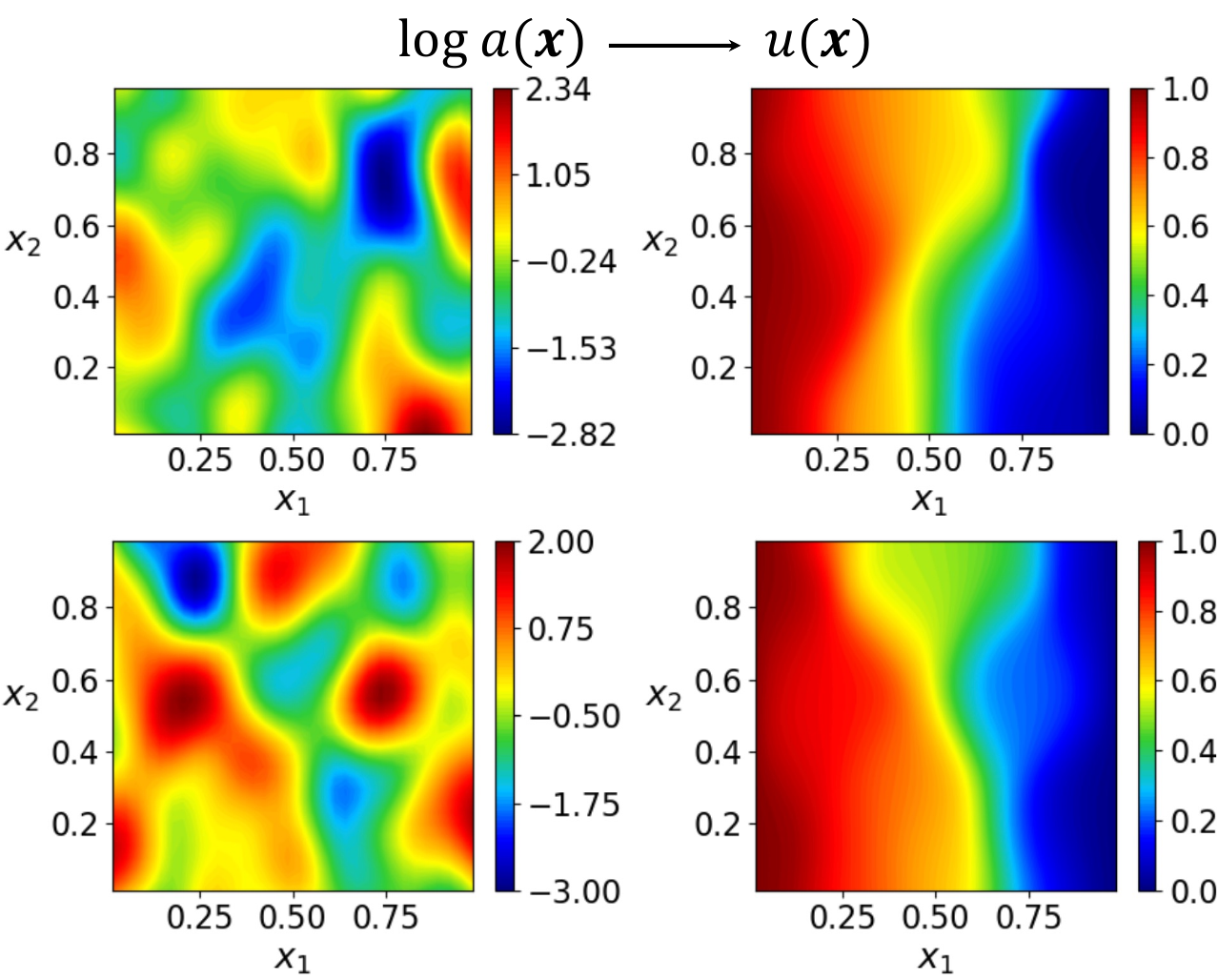}
\end{minipage} \\
\bottomrule
\end{tabular}
\label{tab:examples}
\end{table}

\begin{table}[H]
\centering
\footnotesize
\caption{Architectures of the DeepONet for the benchmarks considered in this study. The Conv2D layers, representing 2D convolution layers, are defined by the number of output filters, kernel size, stride, padding, and activation function. The Average Pooling layers are specified by kernel size, stride, and padding. The ResNet layers, referring to residual networks, are configured with the number of ResNet blocks, the number of layers per block, the number of neurons in each layer, and the activation function. Additionally, MLP refers to multi-layer perceptron.}
\begin{tabular}{p{1.5cm} p{3.5cm} p{3.5cm} p{4.5cm}}
\toprule
\textbf{Case} & \textbf{Dynamical System} & \textbf{Diffusion-reaction} & \textbf{Heat equation} \\ 
\midrule
\textbf{Branch Network} & 
MLP: [100, 40, ReLU, 40, ReLU, 40, ReLU, 40] & 
MLP: [100, 64, ReLU, 64, ReLU, 64, ReLU, 128] & 
\begin{tabular}[c]{@{}l@{}}
Input: (32, 32) \\
Conv2D: (40, (3, 3), 1, 0, ReLU) \\
Average pooling: ((2, 2), 2, 0) \\
Conv2D: (60, (3, 3), 1, 0, ReLU) \\
Average pooling: ((2, 2), 2, 0) \\
Conv2D: (100, (3, 3), 1, 0, ReLU) \\
Average pooling: ((2, 2), 2, 0) \\
Flatten() \\
MLP: [150, ReLU, 150, ReLU, 150]
\end{tabular} \\ 
\midrule
\textbf{Trunk Network} & 
MLP: [1, 40, ReLU, 40, ReLU, 40, ReLU, 40] & 
MLP: [2, 64, ReLU, 64, ReLU, 64, ReLU, 128] & 
\begin{tabular}[c]{@{}l@{}}
MLP: [2, 150, ReLU] \\
ResNet: (2, 2, 150, ReLU) \\
MLP: [150]
\end{tabular} \\ 
\midrule
\textbf{Batch size} & 256 & 256 & 256 \\ 
\midrule
\textbf{Constant Learning Rate} & $1\mathrm{e}{-3}$ & $1\mathrm{e}{-3}$ & $1\mathrm{e}{-4}$ \\ 
\midrule
\textbf{Number of Epochs} & 1000 & 1000 & 20000 \\ 
\bottomrule
\end{tabular}
\label{tab:architectures-hyperparameters}
\end{table}

In this work, the performance of DeepONet is evaluated on the test samples based on two metrics:
\begin{enumerate}
\item The mean $R^2$ score (coefficient of determination), defined as:
\begin{equation}
\label{eqn:mean_$R^2$_score}
\overline{R^2}_{\text{test}} = \frac{1}{N_{\text{test}}} \sum_{i=1}^{N_{\text{test}}} \left(1 - \frac{\sum_{k=1}^{N_{\text{out}}}(u_{\text{truth}, k}^{(i)} - u_{\text{pred}, k}^{(i)})^2}{\sum_{k=1}^{N_{\text{out}}}(u_{\text{truth}, k}^{(i)} - \bar{u}_{\text{truth}}^{(i)})^2}\right),
\end{equation}
where $i$ indexes all test samples, $k$ indexes all the output sensor locations, $u_{\text{truth}, k}^{(i)}$ and $u_{\text{pred}, k}^{(i)}$ are the true and predicted values at the $k^{\text{th}}$ sensor location for the $i^{\text{th}}$ test sample, respectively, and $\bar{u}_{\text{truth}}^{(i)}$ is the mean of $u_{\text{truth}, k}^{(i)}$ for the $i^{\text{th}}$ test sample. 
The mean $R^2$ score measures how well a predictive model captures the variance in the true data across multiple test samples. 
Ranging from $-\infty$ to $1$, an $R^2$ score of $1$ signifies a perfect prediction, $0$ indicates that the model is no better than predicting the mean of the true values, and negative values suggest worse performance than the mean-based model. 
Averaging this score across multiple test samples provides a comprehensive measure of the model’s ability to generalize to new data.
\item The mean squared error (MSE) is defined as:
\begin{equation}
\label{eqn:mean_mse}
\text{MSE}_{\text{test}} = \frac{1}{N_{\text{test}}} \sum_{i=1}^{N_{\text{test}}} \frac{1}{N_{\text{grid}}} \sum_{k=1}^{N_{\text{grid}}} (u_{\text{truth}, k}^{(i)} - u_{\text{pred}, k}^{(i)})^2,
\end{equation}
where $i$ indexes all test samples, $k$ indexes all the output sensor locations, $u_{\text{truth}, k}^{(i)}$ and $u_{\text{pred}, k}^{(i)}$ are the true and predicted values at the $k^{\text{th}}$ sensor location for the $i^{\text{th}}$ test sample, respectively.
\end{enumerate}

We conducted ablation studies to evaluate the performance of our approach on generalization error and training time by varying the amount of labeled data available and the number of evaluation points for the trunk net sampled in each iteration. These ablation results are reported for a constant number of training epochs.
The training for all the experiments carried out in Example 1 was performed on an Apple M3 Max with 36GB of RAM. 
Runs for all experiments in Example 2 runs were conducted on a CPU-based system, where jobs could use up to 32 cores, each with 4GB of memory, potentially sharing resources with other tasks. 
Finally, for Example 3, all runs were executed on a GPU-based system with nodes featuring 4 Nvidia A100 GPUs, each with 40GB of memory. 

\subsection{Example 1: Dynamical System with random external force}
In many physical and engineering applications, systems are influenced by random external forces. 
Understanding how these forces affect the system’s state over time is critical for modeling and predicting system behavior. 
This example considers a simple dynamical system driven by a stochastic input, representing scenarios such as mechanical vibrations, economic models, or climate dynamics, where external influences exhibit randomness.
We now consider the following dynamical system:
\begin{equation}\label{eq:phy_deeponet}
    \begin{split}
    &\frac{du}{dt} = s(t), \\
    & u(0) = 0\text{ and } t\in[0,1],\\
    \end{split}
\end{equation}
where $s(t)$ is a Gaussian random process with zero mean and a covariance function defined by the radial basis function (RBF) with a length scale of $\ell = 0.2$ and variance $1$. Specifically,  $s(t) \sim \text{GP}(0, \text{Cov}(t, t'))$;
The aim here is to learn the mapping between $s(t)$ and $u(t)$ i.e., $\mathcal{G}_{\boldsymbol{\theta}}: s(t) \to u(t)$.

For our experiments, we generated 10,000 pairs of input-output functions, with each function discretized at 100 equally spaced points on the temporal axis.
Using this data, we conducted an ablation study by varying the number of training samples $N_\text{train} \in  \{500, 1000, 2000, 4000, 8000\}$ and the number of evaluation points $N_\text{eval} \in  \{1, 10, 50, 100\}$, across $10$ different seeds (independent trials). The performance of the model was evaluated on $N_\text{test}=2,000$.
These models were trained for $1,000$ epochs following the procedure outlined in Algorithm~\ref{alg:DeepONet_training}.
It is important to note that when $N_\text{eval} = 100$, the training essentially reverts to the traditional method, where the DeepONet is evaluated at $N_\text{out}$ discretized points.

\begin{figure}[H]
\centering
\includegraphics[width=4.5in]{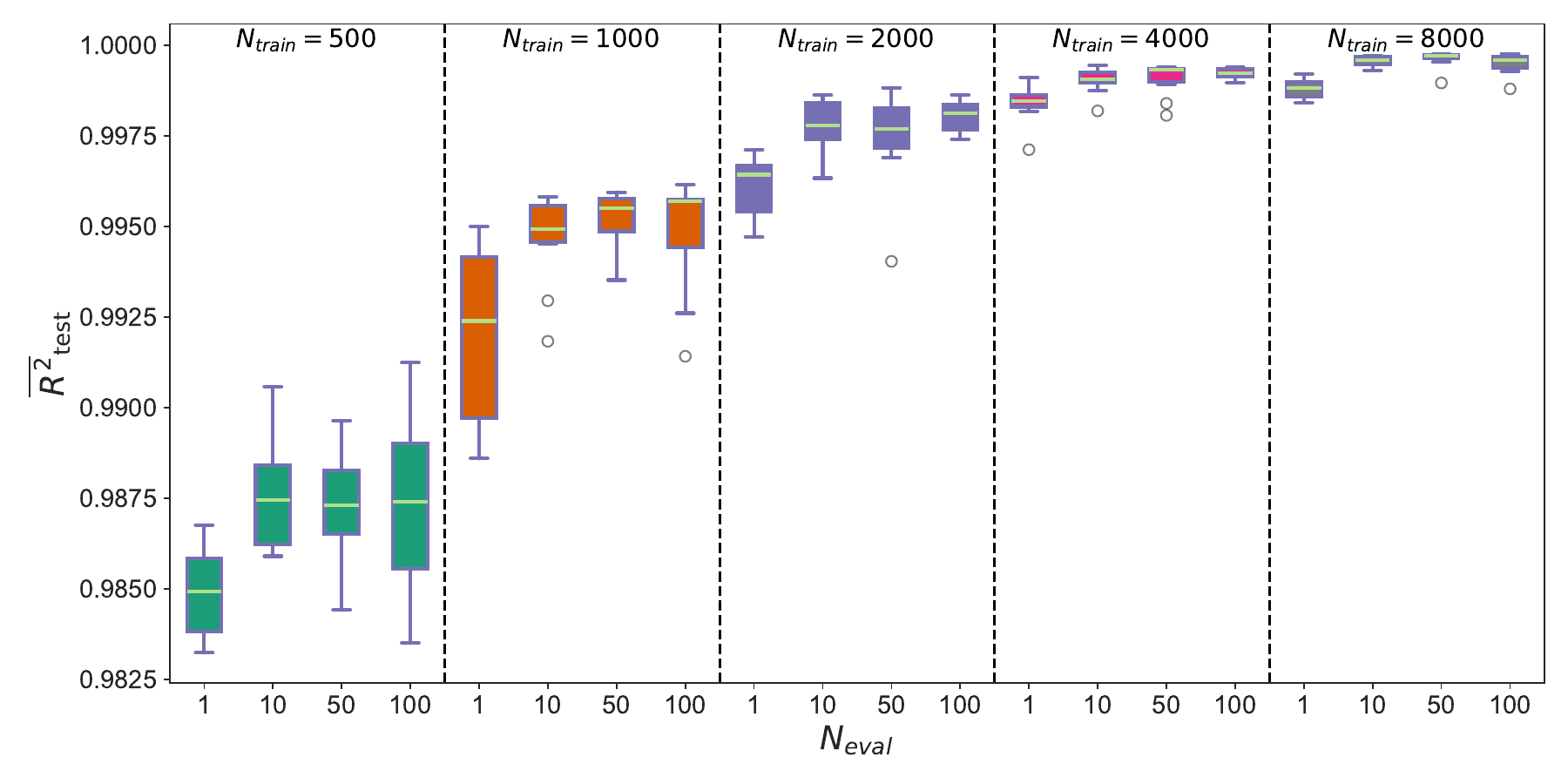}\\
(a)\\[1em] % Adds space between the images
\includegraphics[width=4.5in]{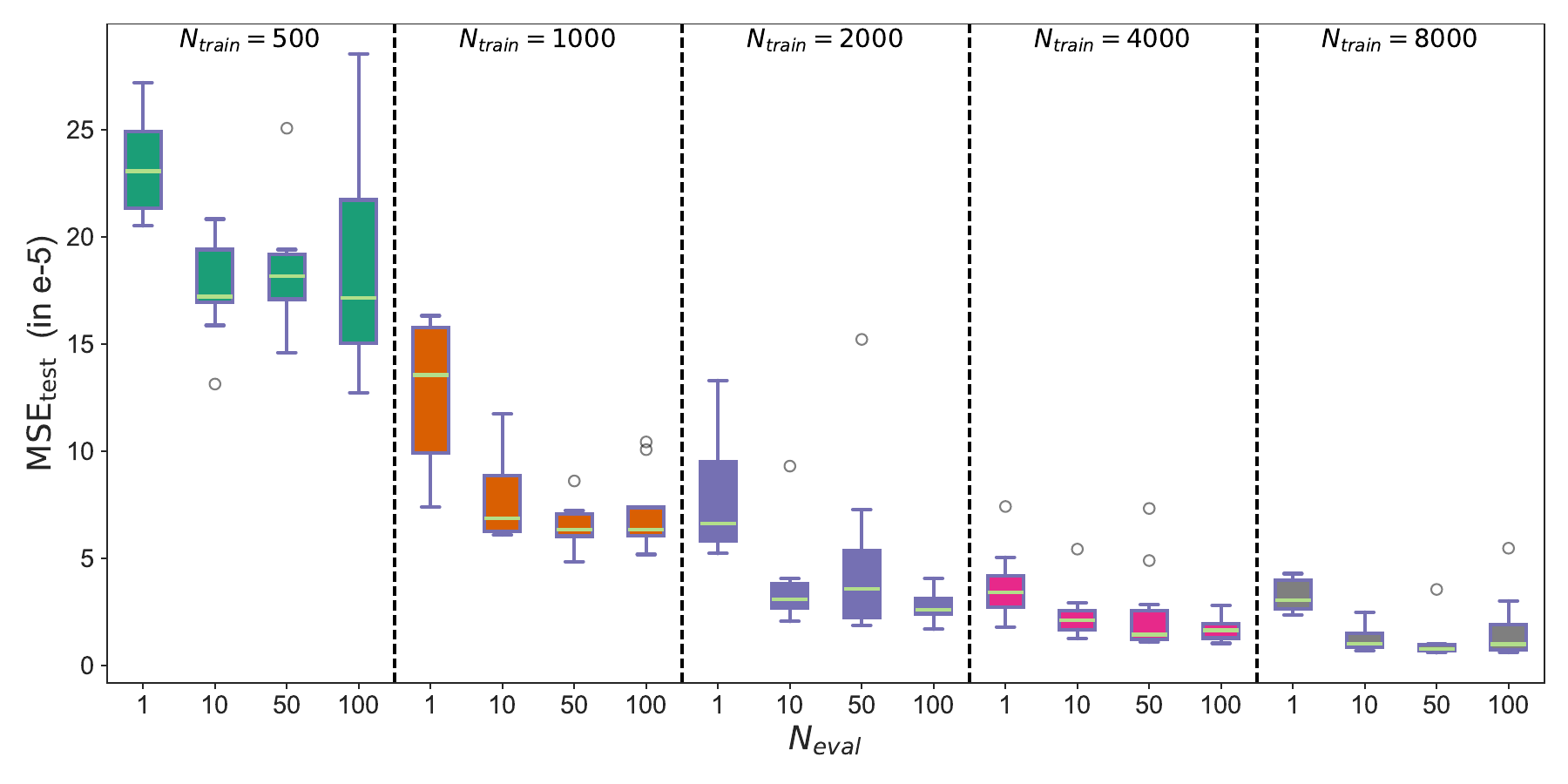}\\
(b)\\[1em] 
\includegraphics[width=4.5in]{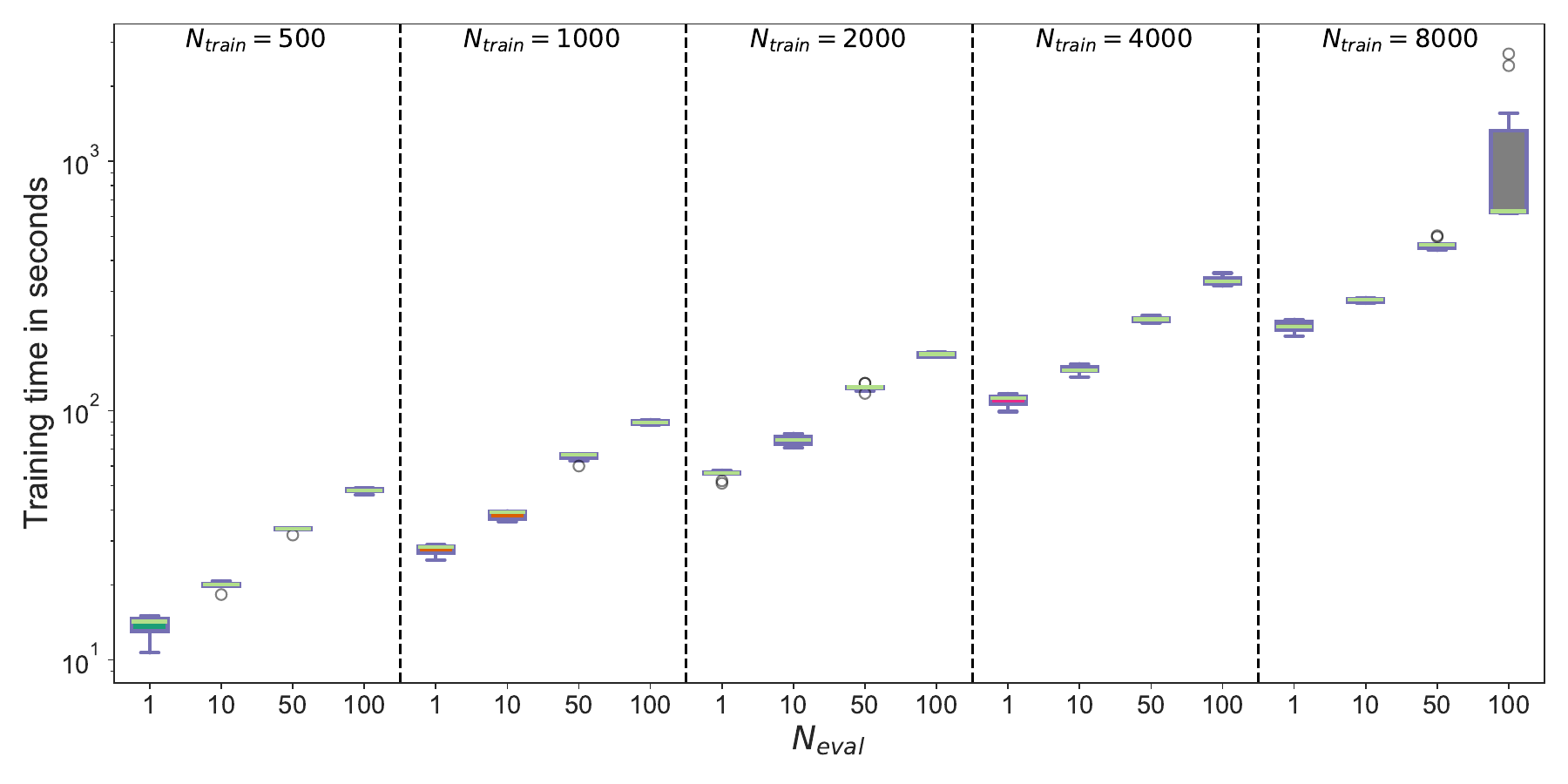}\\
(c)
\caption{For the dynamical system: (a) shows the mean $R^2$ score of the test data, (b) presents the mean squared error of the test data, and (c) displays the training time. The results are based on $10$ independent runs with different seeds, varying the number of training samples and the number of random points at which the output field is evaluated for a given input field sample during training.}
\label{fig:Example1_Boxplots}
\end{figure}

\begin{figure}[H]
\centering
\begin{minipage}{0.8\textwidth}
    \centering
    \includegraphics[width=1.0\textwidth]{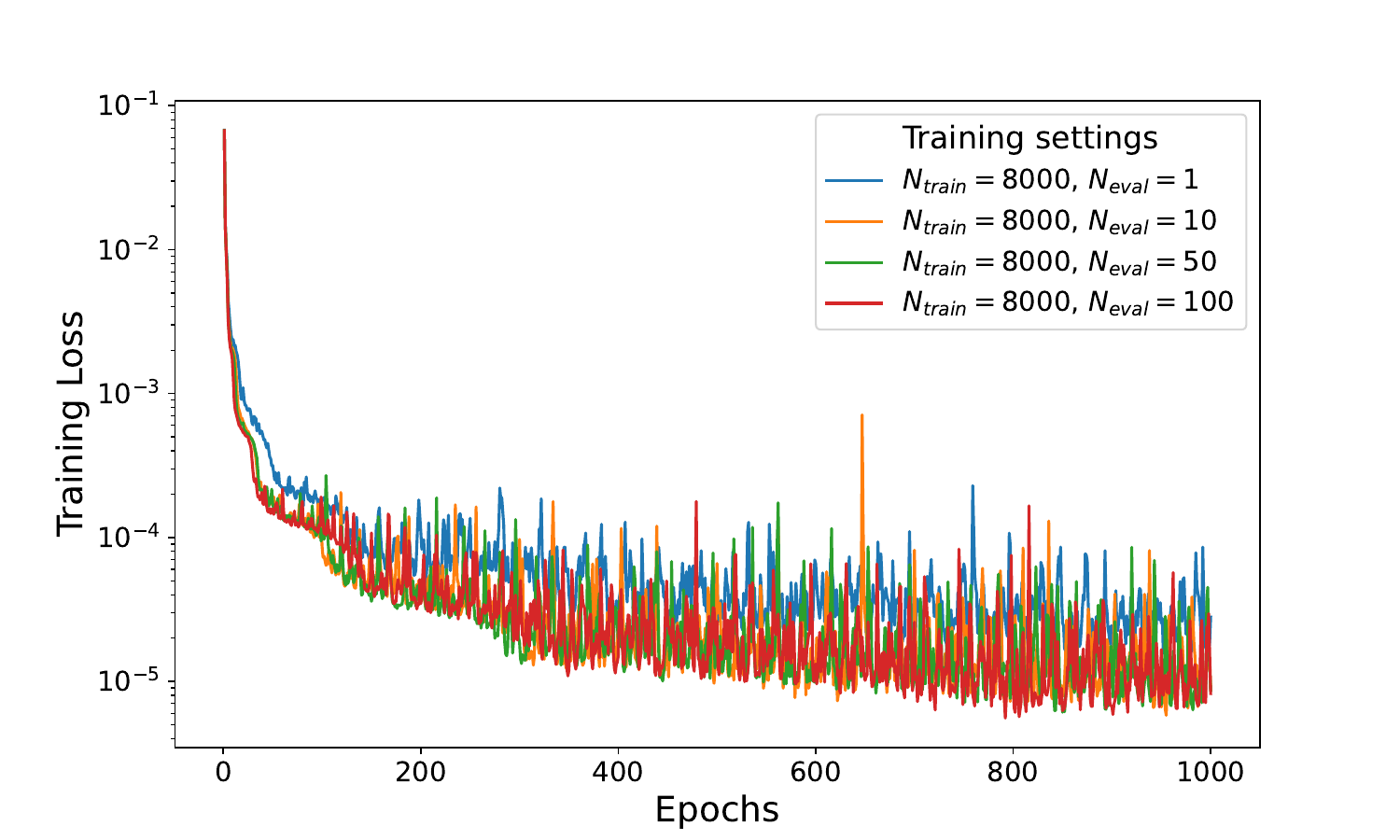}\\
    (a)
\end{minipage}
\\
\begin{minipage}{0.8\textwidth}
    \centering
    \includegraphics[width=1.0\textwidth]{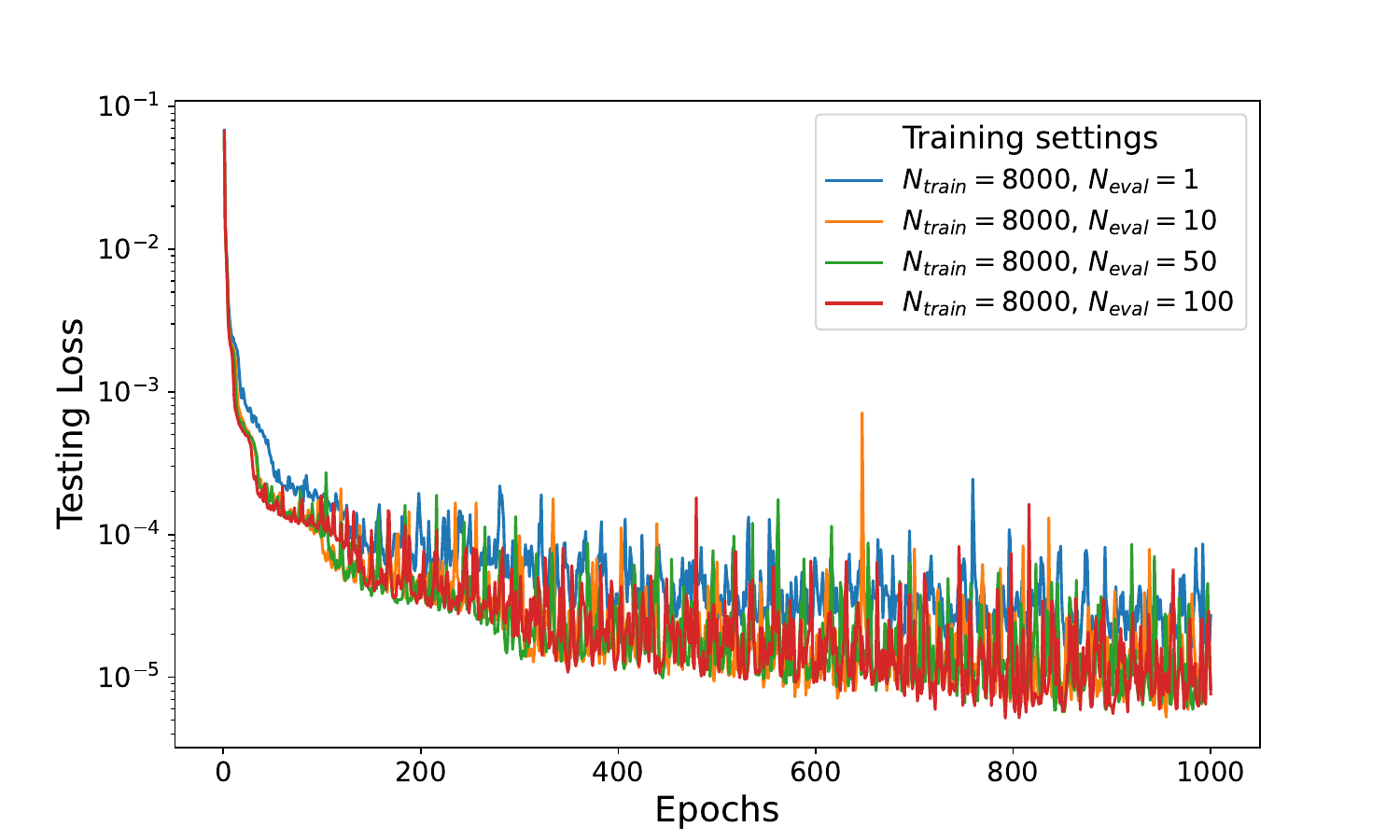}\\
    (b)
\end{minipage}
\caption{For the dynamical system: (a) shows the training loss plot and (b) displays the test loss plot for a specific seed with $N_\text{train} = 8000$ and $N_\text{test} = 2000$, varying the number of sampled points, $N_{\text{eval}}$ used for the trunk network during training.}
\label{fig:Example1_Lossplots}
\end{figure}

\begin{figure}[H]
    \centering
    \includegraphics[width=\textwidth]{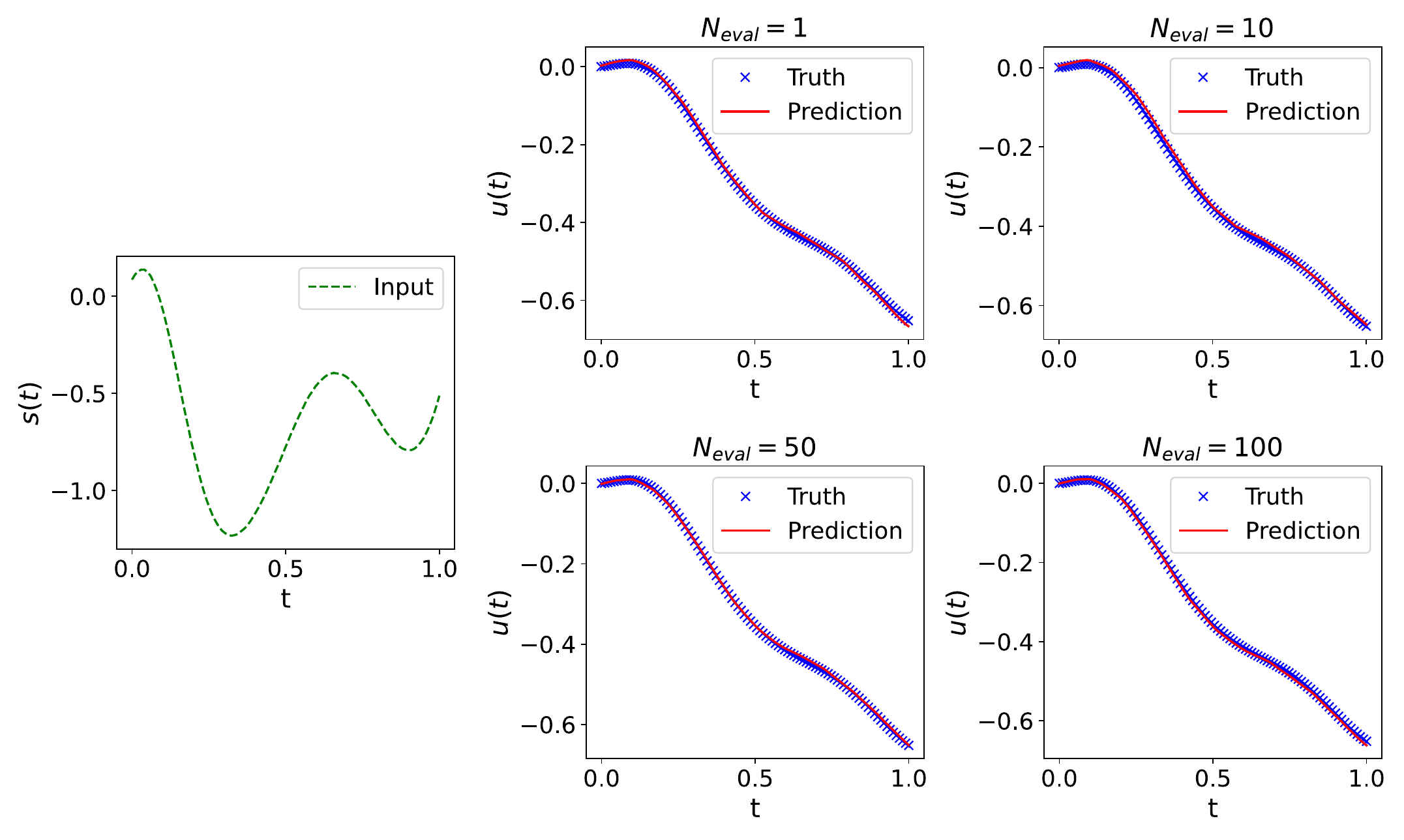}
    \caption{Dynamical system: Comparison of all models for a representative test sample across different $N_\text{eval}$ values, with $N_\text{train} = 2000$.}
    \label{fig:Example1_sample_realization}
\end{figure}

Figure~\ref{fig:Example1_Boxplots} illustrates the box plots of the mean $R^2$ score and the MSE of the test data across different experiments carried out during the ablation study, along with their corresponding computational training times.
As expected, Figure~\ref{fig:Example1_Boxplots}(a) demonstrates that the mean $R^2$ score improves as the number of training samples increases, while the MSE of the test data decreases accordingly. 
Moreover, we can evidently see that, for a given number of training samples, using $N_\text{eval} = 10$ or $50$ evaluation points as input to the trunk network yields comparable or even better accuracy than using $N_\text{eval} = 100$ (which is equal to $N_\text{out}$), with significantly reduced training times.
Additionally, the training and test loss plots in Figure~\ref{fig:Example1_Lossplots} show that training with a randomized trunk using $N_\text{eval} = 10$ or $50$ exhibits slightly higher variance. 
This increased variance helps explore the parameter space more effectively and results in a lower overall loss.
Conversely, when $N_\text{eval} = 1$, the accuracy metrics ($\overline{R^2}_{\text{test}}$ and $\text{MSE}_{\text{test}}$) are notably poor, as the model does not encounter enough data in each iteration. 
These results substantiate our hypothesis that randomizing the inputs to the trunk network effectively enhances the training process and markedly reduces training time.
In Figure~\ref{fig:Example1_sample_realization}, a comparison of all models for a representative test sample across different $N_\text{eval}$ values is shown for $N_\text{train} = 2000$.

Furthermore, we conducted an experiment using $N_{\text{eval}} = 5$ and $N_{\text{eval}} = 10$, comparing its selection to be random and uniformly spaced. We observe that the models utilizing randomly sampled points achieve lower test errors compared to those using uniformly spaced points. 
This improvement is attributed to the model's exposure to diverse information about the function in each iteration due to the randomization. 
These models were trained for $500$ epochs. 

\begin{figure}[H]
    \centering
    \includegraphics[width=\textwidth]{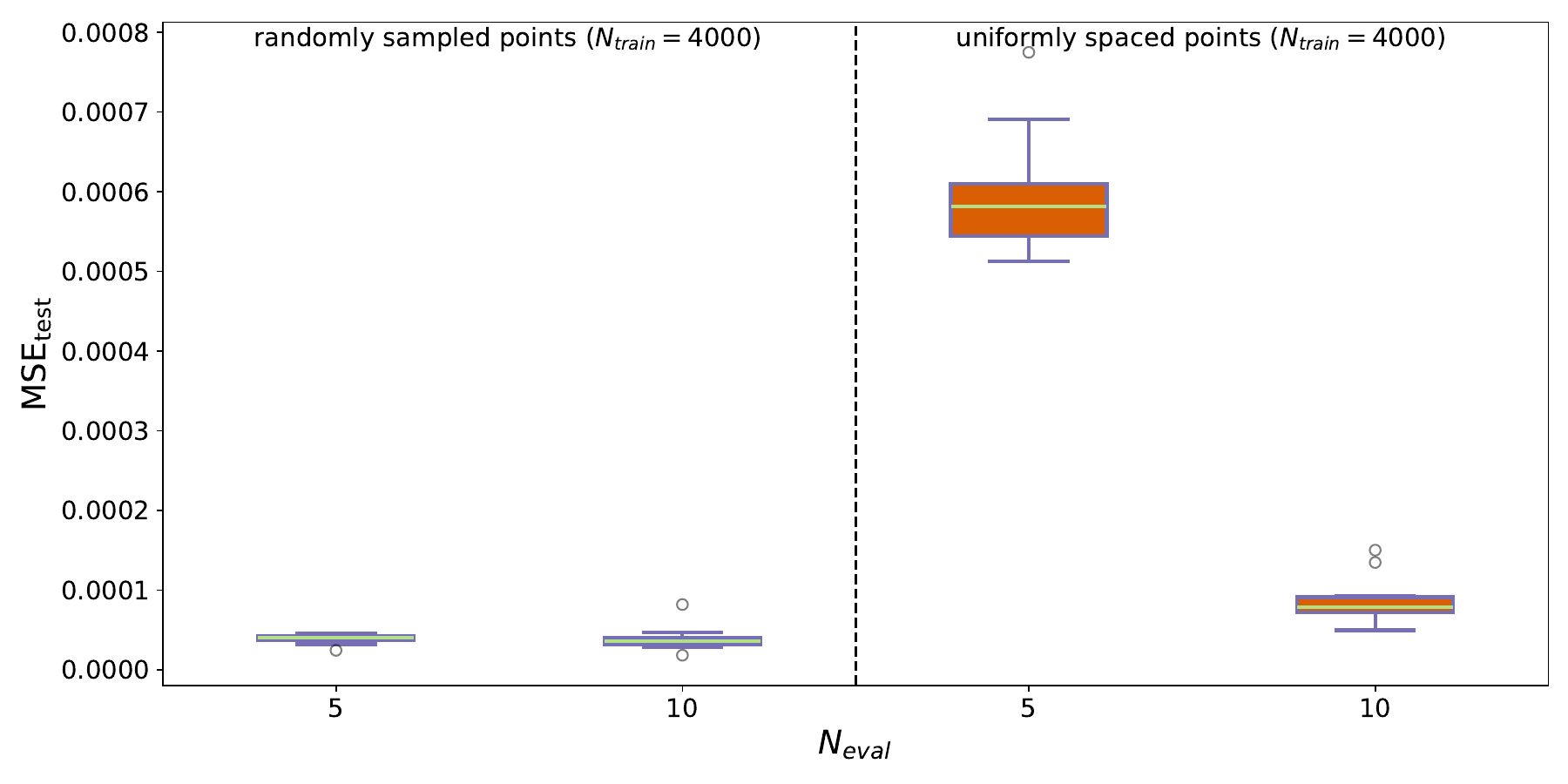}
    \caption{For the dynamical system: mean squared error of the test data using $5$ and $10$ evaluation points, both randomly sampled and uniformly spaced, as inputs to the trunk network, with $N_\text{train} = 4000$.}
    \label{fig:Example1_Study}
\end{figure}

\subsection{Example 2: Diffusion-reaction dynamics with uncertain source field}
Diffusion-reaction systems model a wide range of physical, chemical, and biological processes, such as heat conduction with internal heat generation, chemical reactions in porous media, and population dynamics in ecology. 
In this example, we simulate a one-dimensional diffusion-reaction process, which can represent phenomena like pollutant dispersion in soil or chemical concentration in a reactor. 
Understanding how the concentration  $u(t,x)$  evolves over time and space under random sources  $s(x)$  is crucial for predicting and optimizing such processes.
The system is governed by the following equation:
% In this example, we consider a diffusion-reaction system governed by the following equation:
\begin{equation}
\label{eqn:diffusion-reaction}
\begin{aligned}[b]
&\frac{\partial u}{\partial t} = D \frac{\partial^2 u}{\partial x^2} + k u^2 + s(x), \quad (t, x) \in (0, 1] \times (0, 1], \\
&u(0, x) = 0 \ \forall \ x \in (0,1), \\
&u(t, 0) = 0 \ \forall \ t \in (0,1), \\
&u(t, 1) = 0 \ \forall \ t \in (0,1),
\end{aligned}
\end{equation}
where $D = 0.01$ is the diffusion coefficient and $k = 0.01$ is the reaction coefficient. The source term $s(x)$ is modeled as a random field generated from a Gaussian random process with zero mean and radial basis covariance function characterized by a length scale of $\ell = 0.2$ and variance $1$. 
The goal is to learn the solution operator that maps these random source terms $s(x)$ to their corresponding solutions $u(t, x)$, i.e., $\mathcal{G}_{\bm{\theta}}: s(x) \to u(t, x)$.
For our studies, we generated $2,500$ pairs of input source field and output solution field functions. 
Each source function is discretized at $100$ equally spaced points in space, while the solution field is discretized at $101$ points in time and $100$ points in space, resulting in $101 \times 100$ grid points.

\begin{figure}[H]
\centering
\includegraphics[width=4.5in]{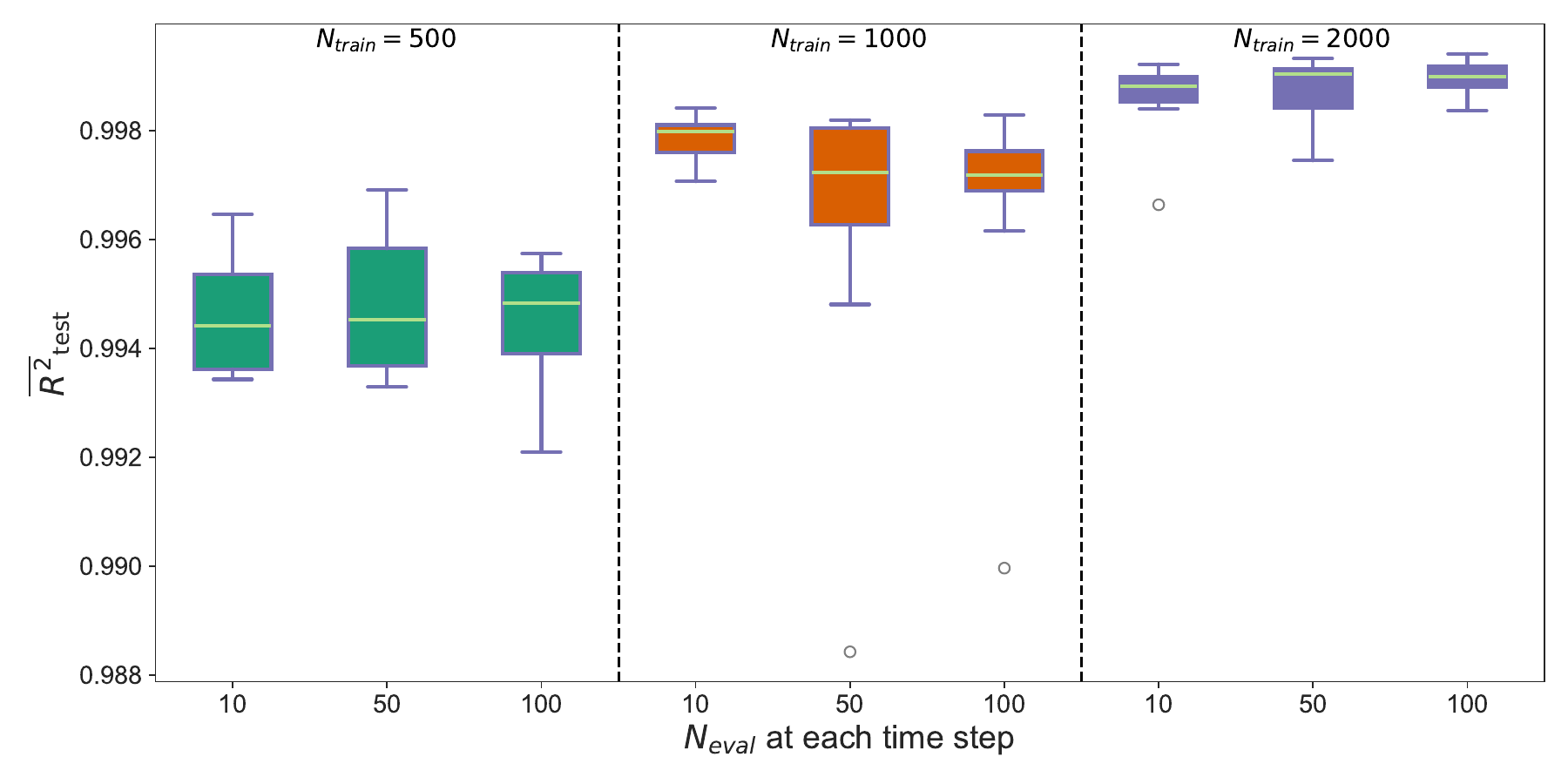}\\
(a)\\[1em] % Adds space between the images
\includegraphics[width=4.5in]{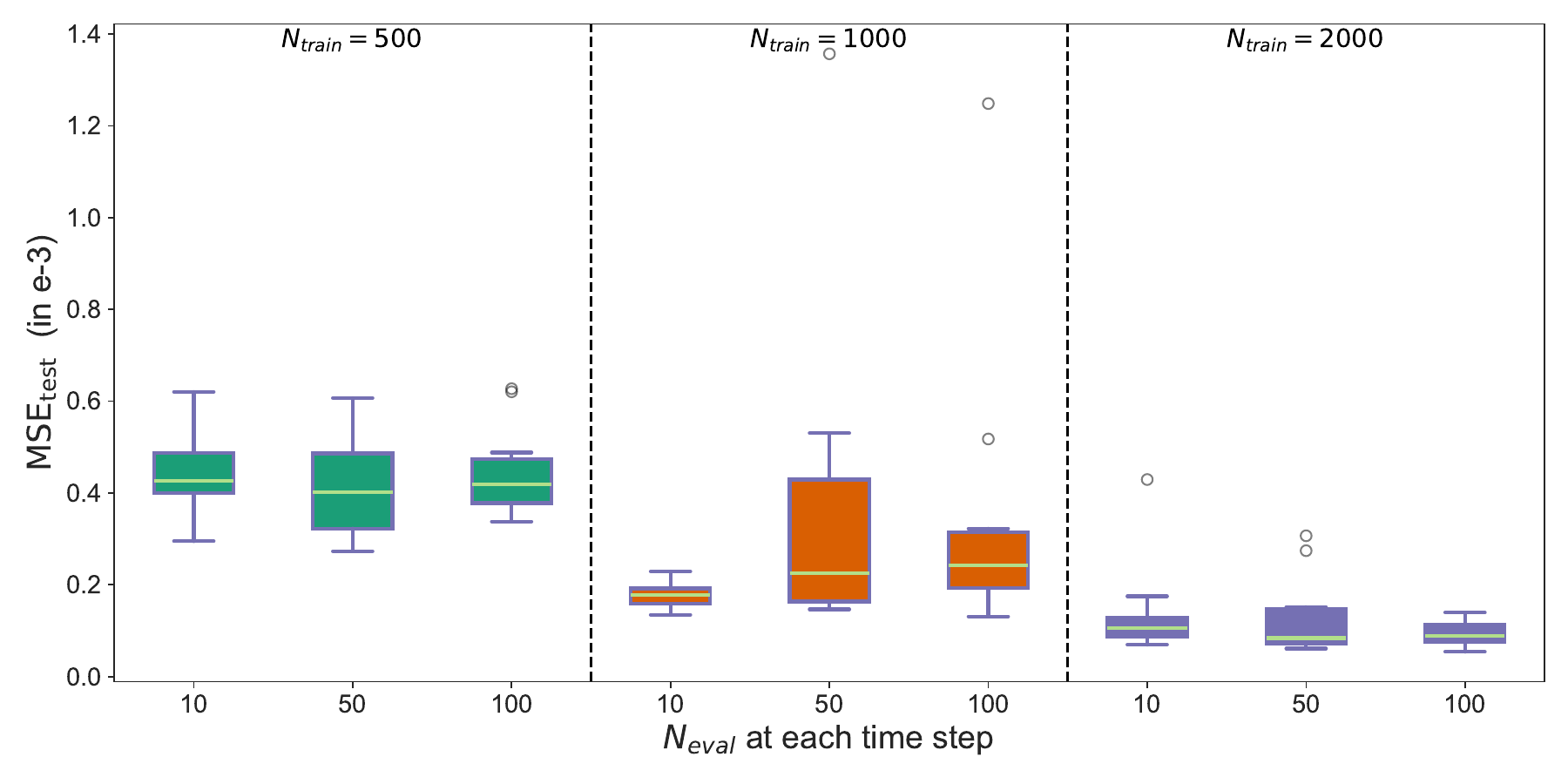}\\
(b)\\[1em]
\includegraphics[width=4.5in]{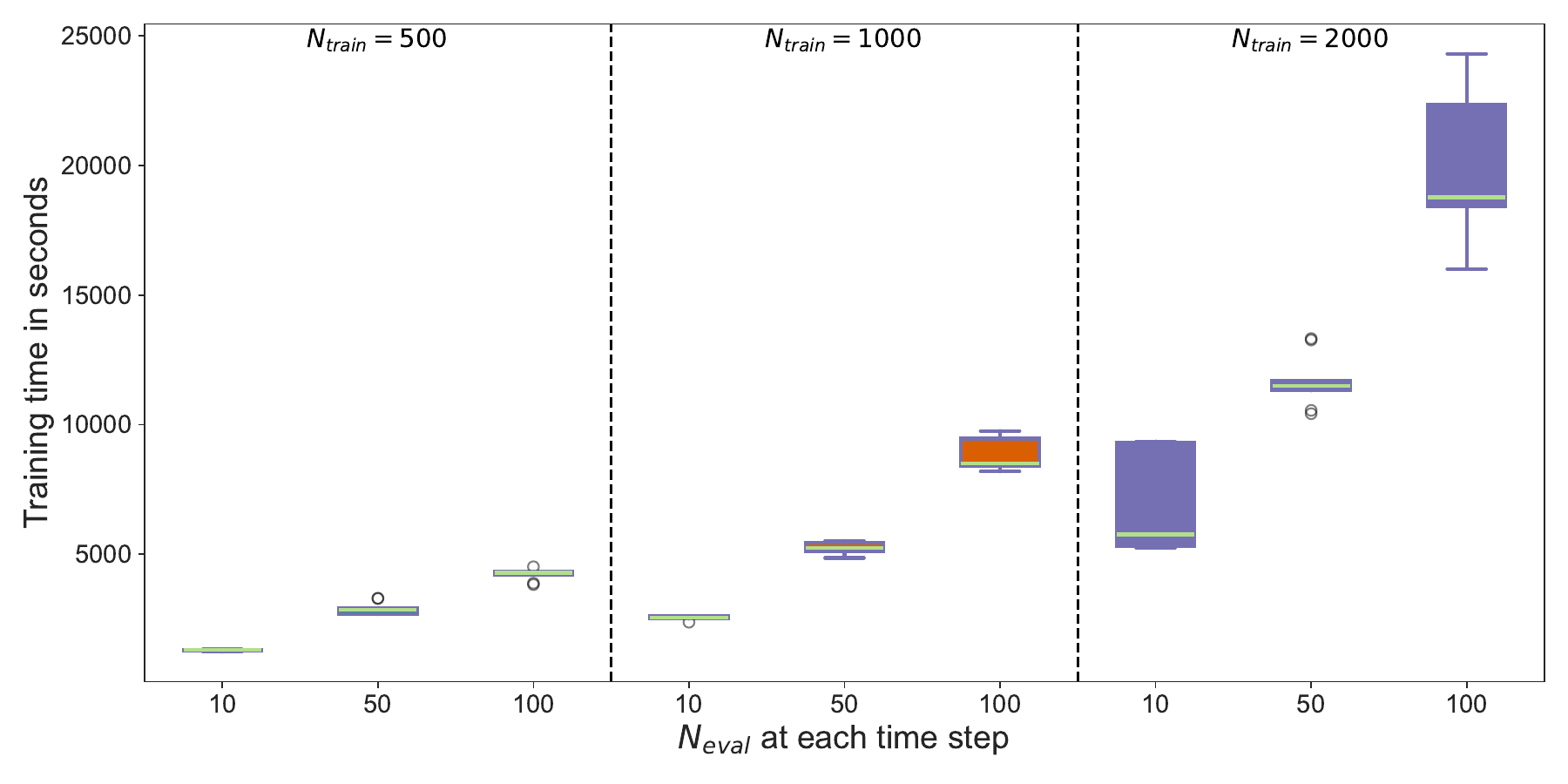}\\
(c)
\caption{For the diffusion-reaction dynamics: (a) shows the mean $R^2$ score of the test data, (b) presents the mean squared error of the test data, and (c) displays the training time. The results are based on $10$ runs with different seeds, varying the number of training samples and the number of evaluation points at each time step for the trunk network.}
\label{fig:Example2_Boxplots}
\end{figure}

\begin{figure}[H]
\centering
\begin{minipage}{0.8\textwidth}
    \centering
    \includegraphics[width=1.0\textwidth]{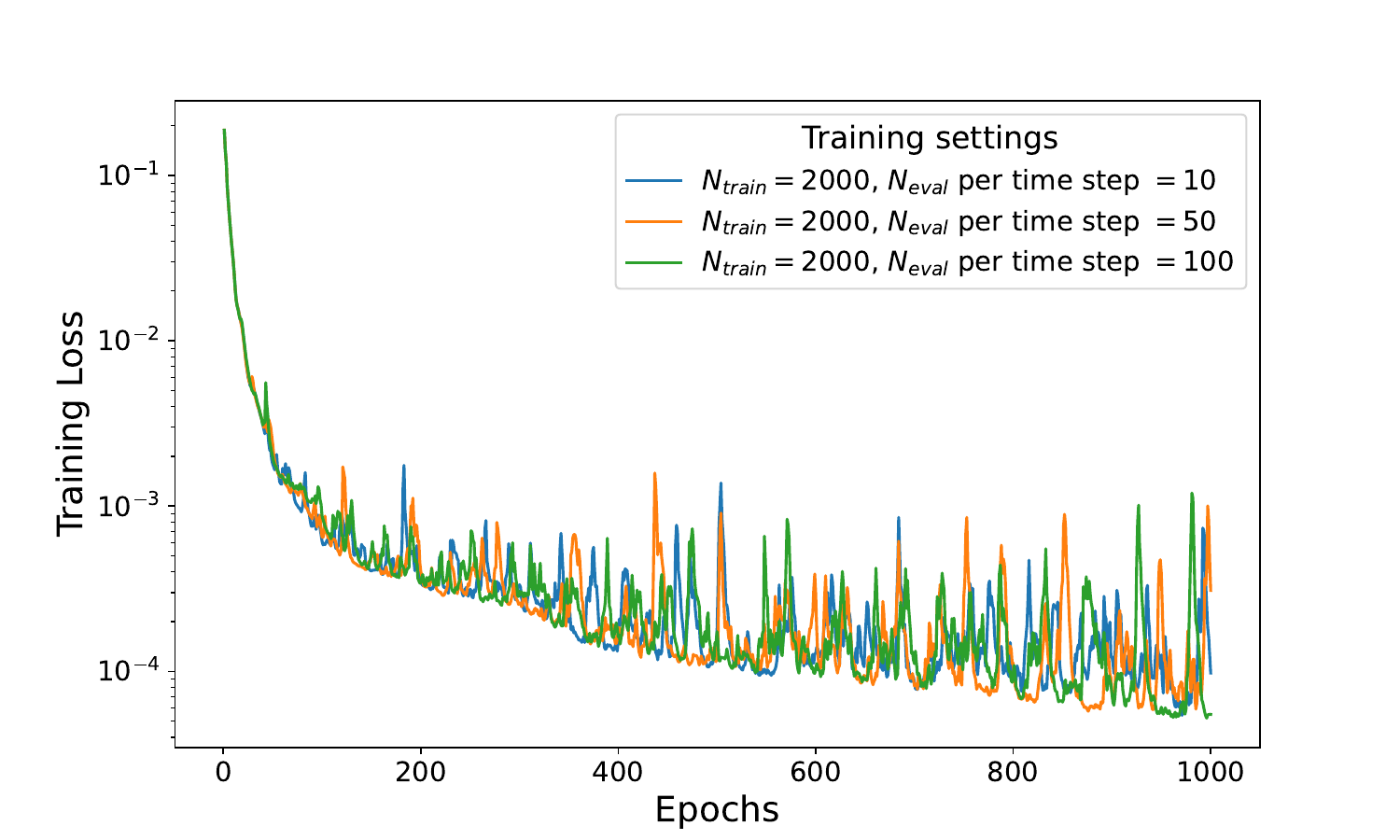}\\
    (a)
\end{minipage}
\\
\begin{minipage}{0.8\textwidth}
    \centering
    \includegraphics[width=1.0\textwidth]{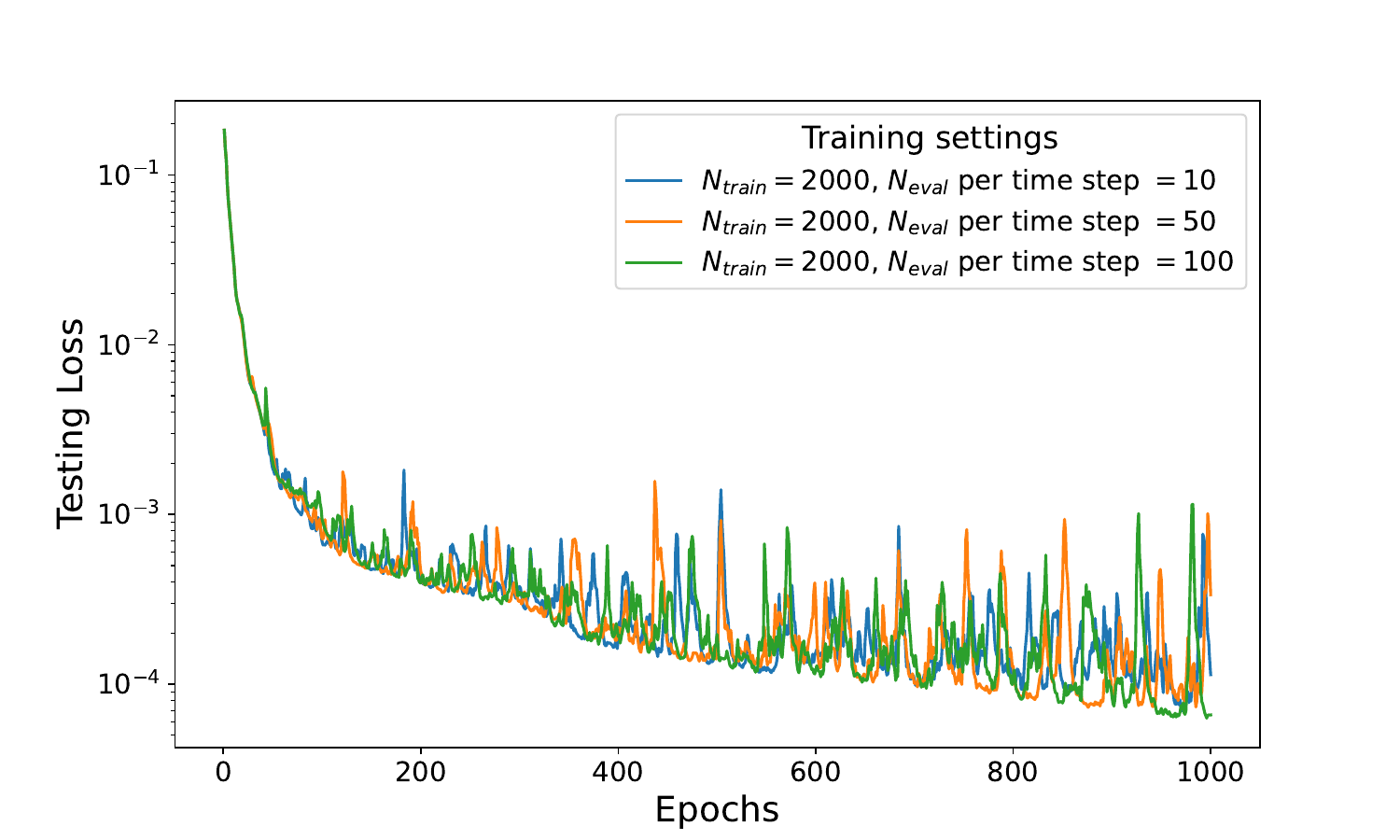}\\
    (b)
\end{minipage}
\caption{For the diffusion-reaction dynamics: (a) shows the training loss plot and (b) displays the test loss plot for a specific seed with $N_\text{train} = 2000$, varying the number of evaluation points used for the trunk network per time step.}
\label{fig:Example2_Lossplots}
\end{figure}

\begin{figure}[H]
    \centering
    \includegraphics[width=\textwidth]{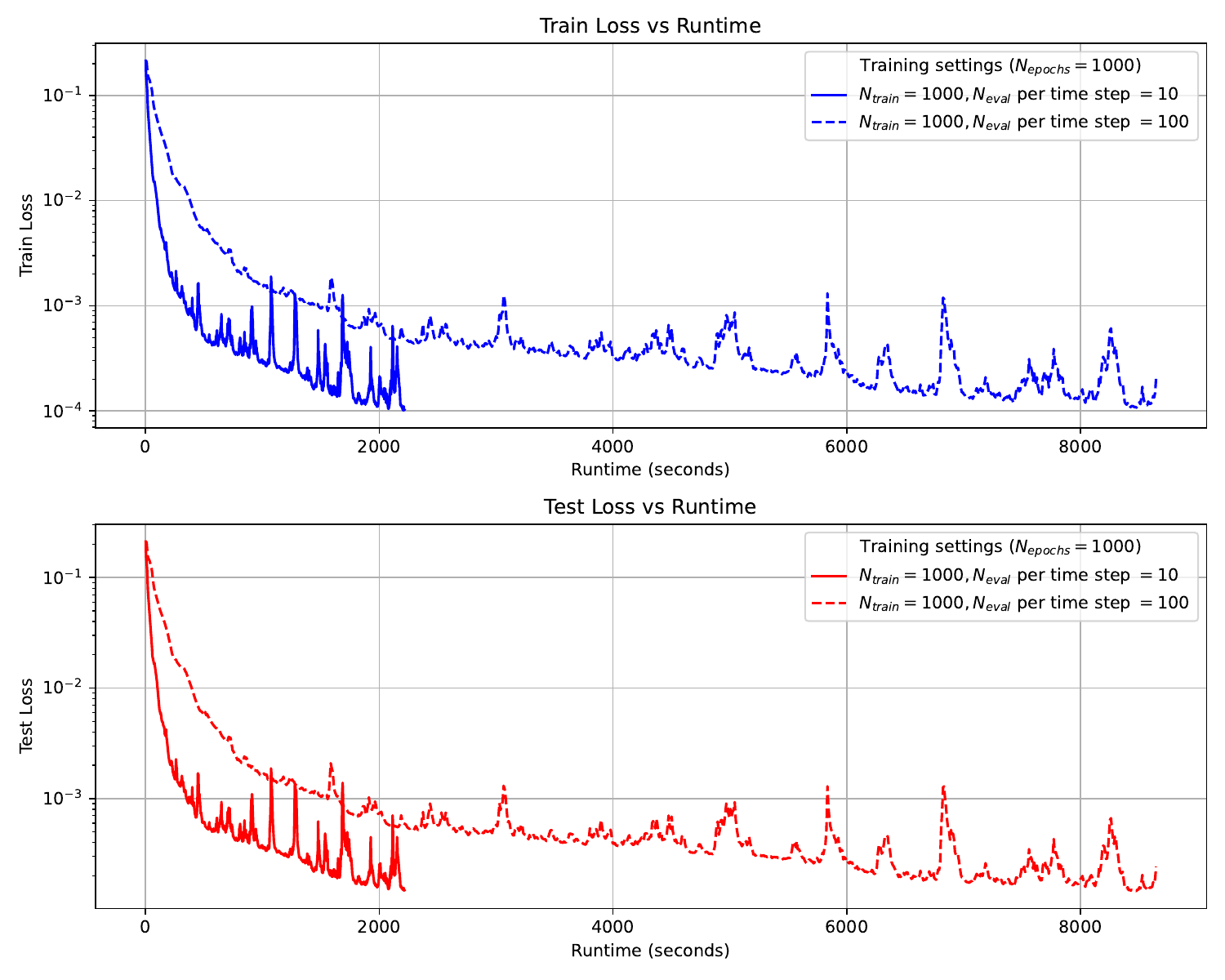}
    \caption{Diffusion-reaction dynamics: Comparison of the train and test losses with respect to runtime.} 
    \label{fig:Example2_train_test_loss_vs_runtime}
\end{figure}

\begin{figure}[H]
    \centering
    \includegraphics[width=\textwidth]{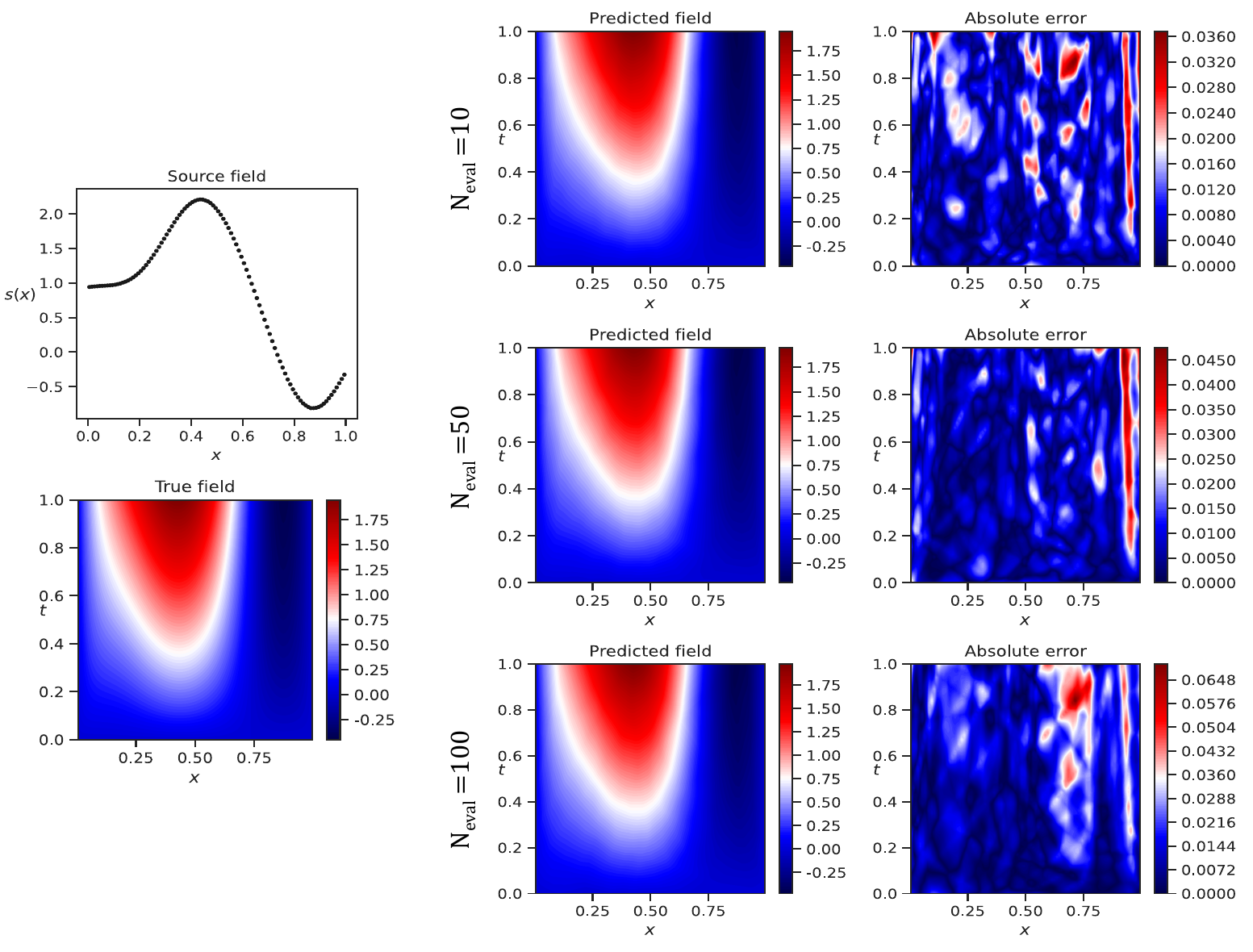}
    \caption{Diffusion-reaction dynamics: Comparison of all models for a representative test sample across different $N_\text{eval}$ values per time step, with $N_\text{train} = 1000$.} 
    \label{fig:Example2_sample_realization}
\end{figure}

In line with the previous examples, we conducted ablation studies by varying both the number of training samples, $N_\text{train} \in \{500, 1000, 2000\}$, and the number of spatial evaluation points used at each time step per iteration as $N_\text{eval} \in \{10, 50, 100\}$. These experiments were performed across $10$ different random seeds, with a test set comprising $N_\text{test} = 500$ samples, following the procedure described in Algorithm~\ref{alg:DeepONet_training}. Each model was trained for $1000$ epochs. 
As observed in Figure~\ref{fig:Example2_Boxplots}(a), the model performance with $N_\text{eval}$ values of $10$ and $50$ per time step is comparable to, or better than, the performance with $N_\text{eval} = 100$ per time step, which corresponds to evaluating the trunk network at all available evaluation points.
This strengthens our hypothesis that randomizing the inputs to the trunk network effectively improves the training process, while the training time is drastically reduced (see Figure~\ref{fig:Example2_Boxplots}(c)). 
Furthermore, the training and test loss plots in Figure~\ref{fig:Example2_Lossplots} show that training with a randomized trunk using $N_\text{eval} = 10$ or $50$ per time step exhibits similar behavior as $N_\text{eval} = 100$. Additionally, Figure~\ref{fig:Example2_train_test_loss_vs_runtime} illustrates that the train and test losses associated with our randomized approach decrease at a significantly faster rate than those of the standard approach. The randomized method not only converges more rapidly but also achieves lower loss values within a shorter runtime, underscoring its efficiency and effectiveness in enhancing convergence.
In Figure~\ref{fig:Example2_sample_realization}, a comparison of all models for a representative test sample across different $N_\text{eval}$ values per time step is shown for $N_\text{train} = 1000$.

\subsection{Example 3: Heat equation with uncertain conductivity field}
In many engineering and physics applications, the behavior of heat distribution across a medium is modeled using the heat equation. 
This example investigates a scenario where the heat conduction is influenced by spatially varying properties, such as the conductivity of the material, which may vary randomly due to imperfections or other factors. 
The steady-state heat equation models this situation, particularly when the system has reached thermal equilibrium. The scenario might represent problems in materials science, mechanical engineering, or environmental modeling, where temperature distribution is affected by spatially random properties like material conductivity.
We consider the steady-state heat equation, given by:
% In the final example, we investigate the steady-state heat equation given by:
\begin{equation}
\label{eqn:heat-eqn}
\begin{aligned}
&-\nabla \cdot (a(\bx) \nabla u(\bx)) = 0, \\
& \bx = (x_1, x_2), \quad \bx \in \Omega = [0, 1]^2, \\
&u(0, x_2) = 1, \quad u(1, x_2) = 0, \\
&\frac{\partial u(x_1, 0)}{\partial n} = \frac{\partial u(x_1, 1)}{\partial n} = 0,
\end{aligned}
\end{equation}
where $a(\bx)$ denotes spatially varying conductivity field and $u(\bx)$ represents the temperature field.  
The logarithm of the conductivity is modeled as a random field generated from a Gaussian process with zero mean and a radial basis covariance function. 
This function is characterized by length scales of $0.1$ and $0.15$ in the $x_1$ and $x_2$ directions, respectively, with a variance of $1$.
The objective is to learn the solution operator that maps these random conductivity fields $a(\bx)$ to their corresponding temperature distributions $u(\bx)$, i.e., $\mathcal{G}_{\boldsymbol{\theta}}: a(\bx) \mapsto u(\bx)$.

For our study, we generated $5,500$ pairs of input conductivity fields and output temperature fields. 
Each field was discretized into $32 \times 32 = 1024$ spatial points. 
In line with previous examples, we performed ablation studies by varying both the number of training samples, $N_\text{train} \in \{500, 2500, 5000\}$, and the number of evaluation points, $N_\text{eval} \in \{16, 512, 1024\}$ per iteration. These experiments were conducted across $10$ different random seeds, with a test set comprising $N_\text{test} = 500$ samples, following the procedure described in Algorithm~\ref{alg:DeepONet_training}. Each model was trained for $20,000$ epochs.

\begin{figure}[H]
\centering
\includegraphics[width=4.5in]{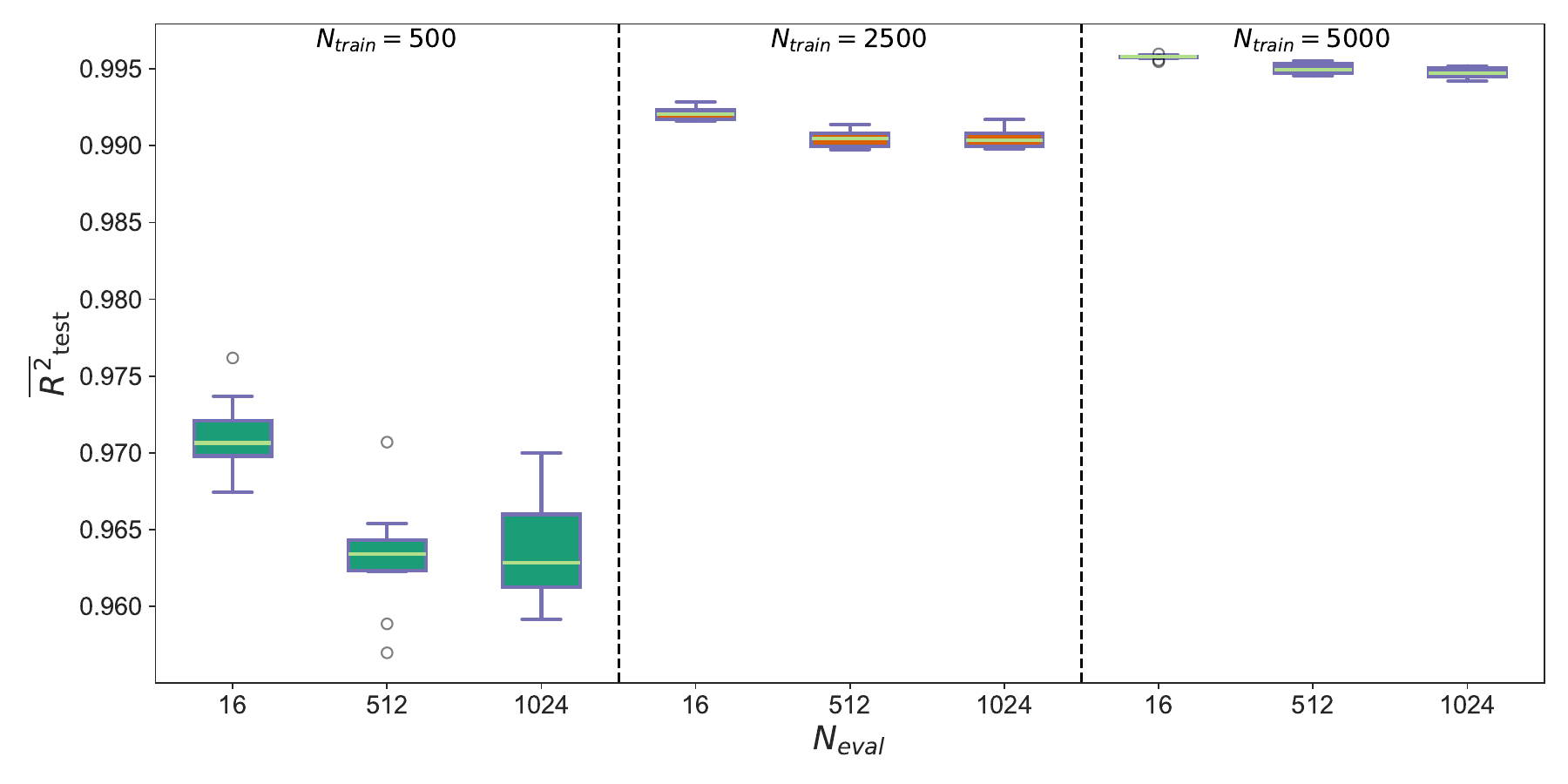}\\
(a)\\[1em] % Adds space between the images
\includegraphics[width=4.5in]{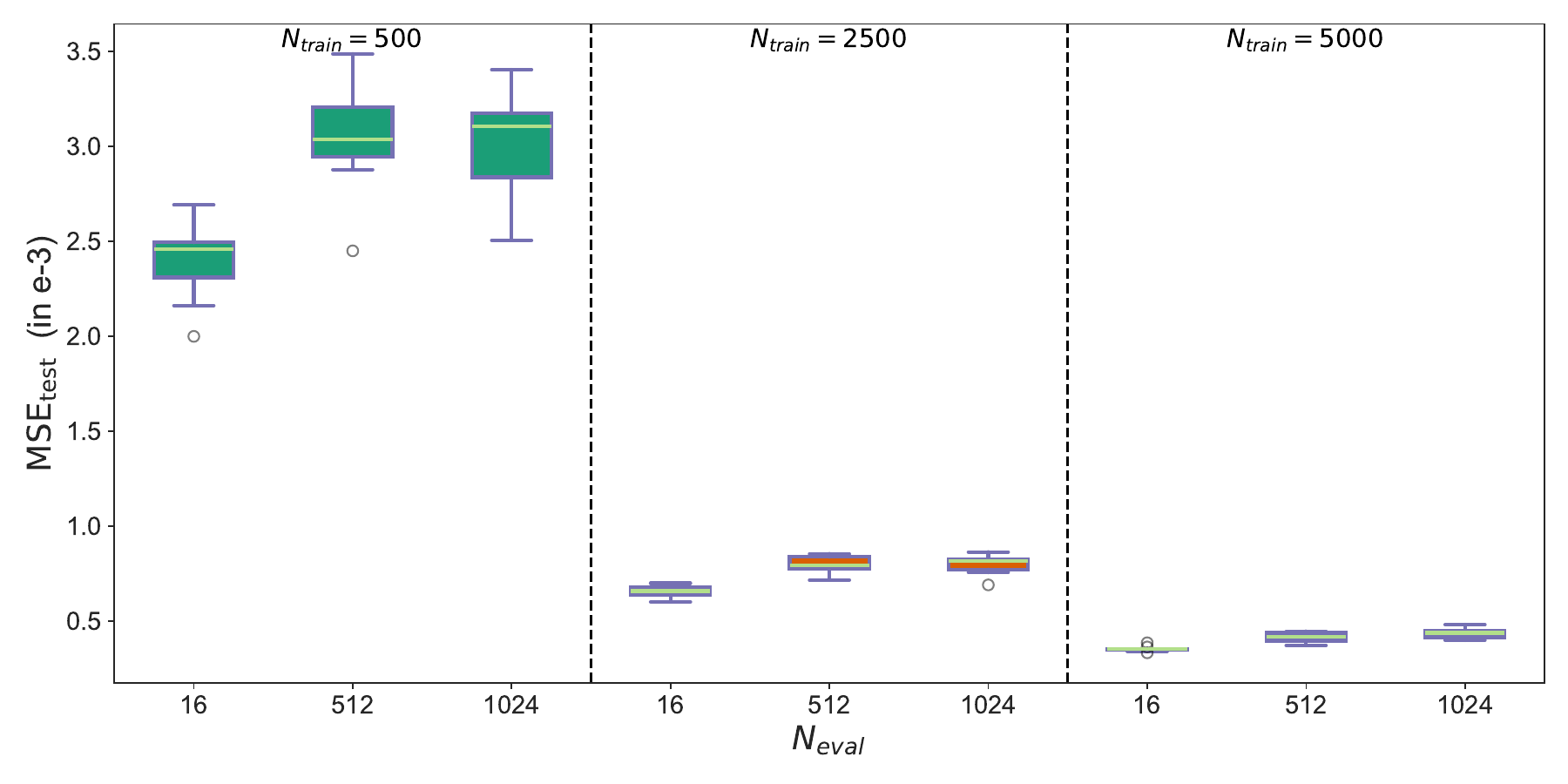}\\
(b)\\[1em]
\includegraphics[width=4.5in]{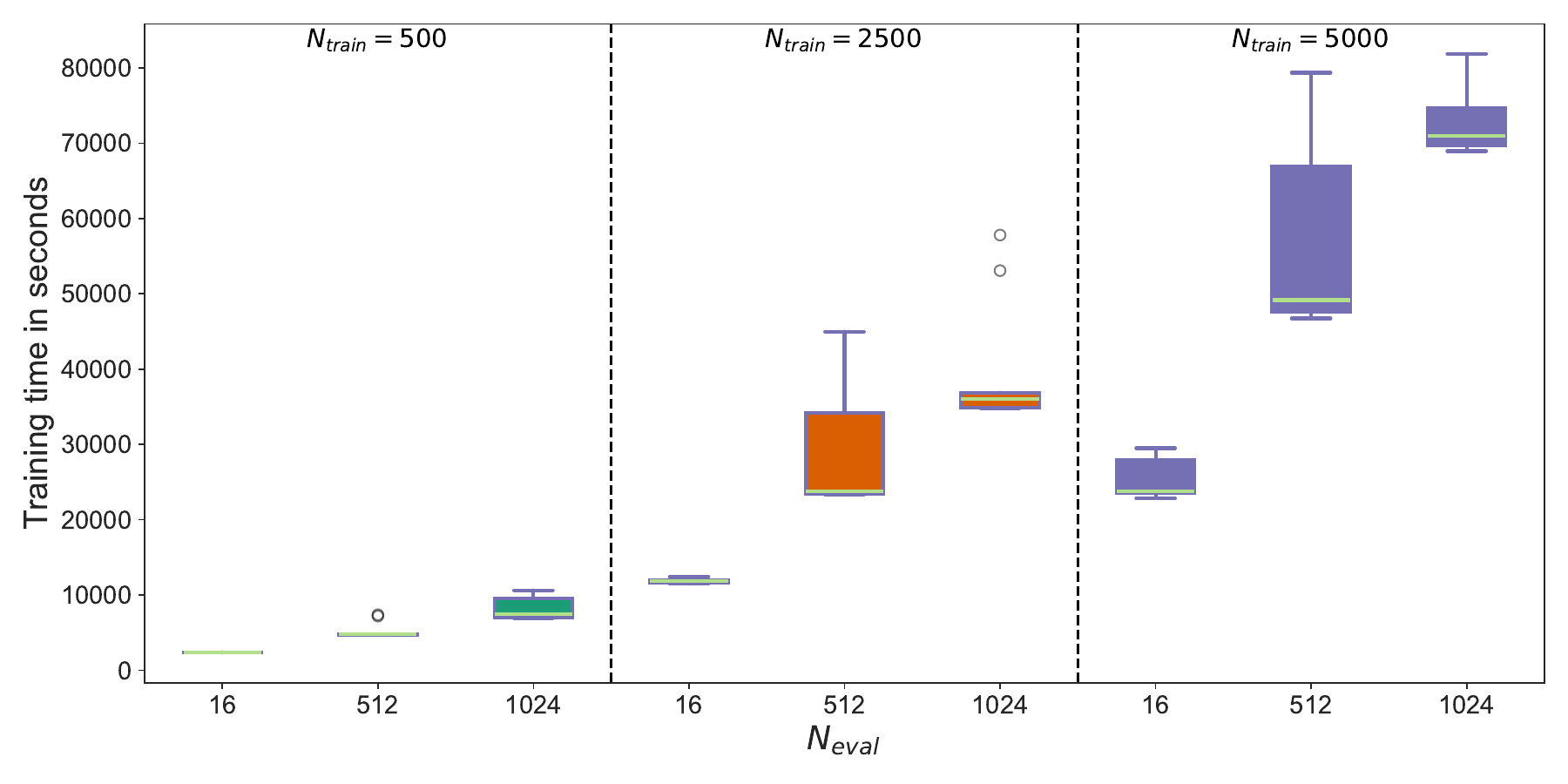}\\
(c)
\caption{For the heat equation: (a) shows the mean $R^2$ score of the test data, (b) presents the mean squared error of the test data, and (c) displays the training time. The results are based on $10$ runs with different seeds, varying the number of training samples and evaluation points for the trunk network.}
\label{fig:Example3_Boxplots}
\end{figure}

\begin{figure}[H]
\centering
\begin{minipage}{0.8\textwidth}
    \centering
    \includegraphics[width=1.0\textwidth]{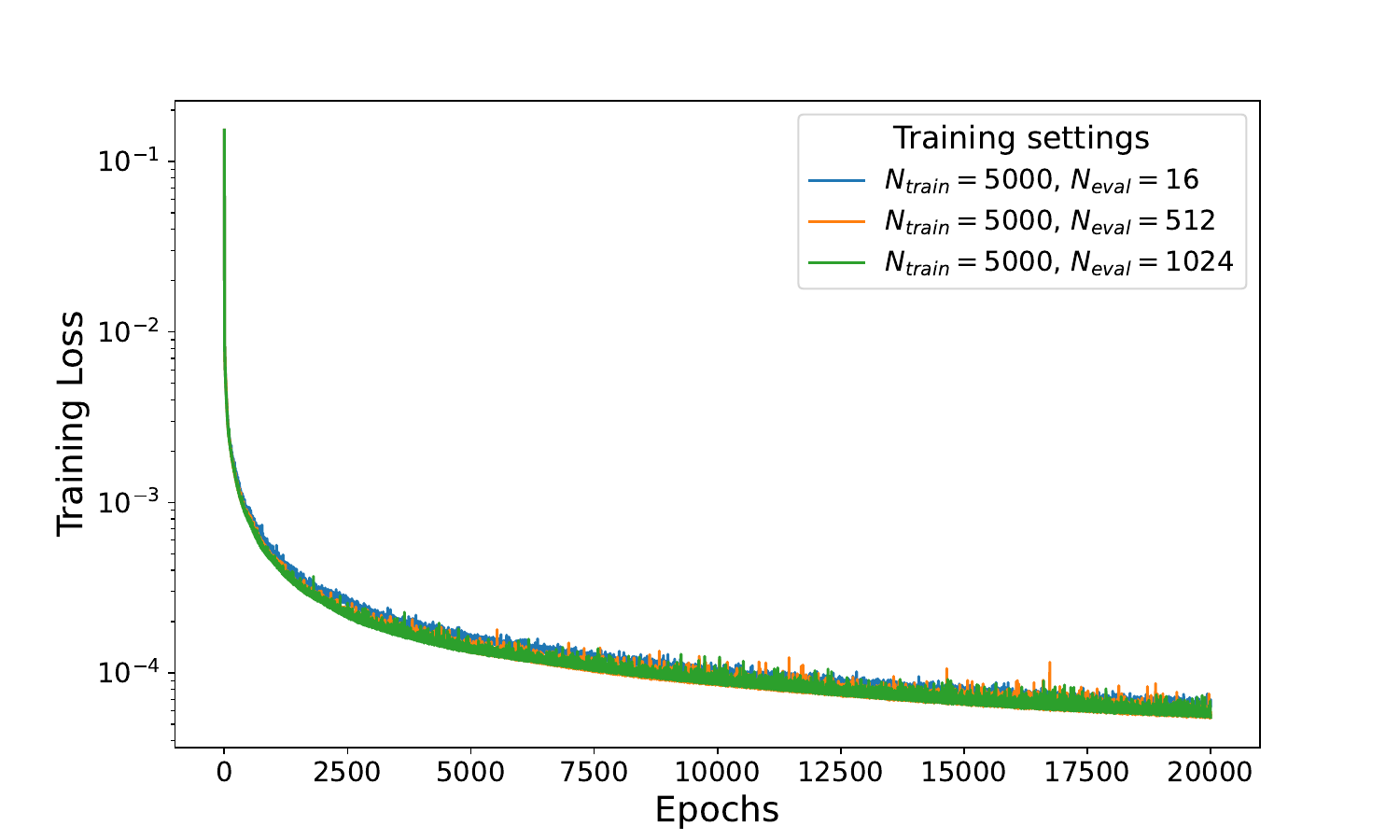}\\
    (a)
\end{minipage}
\\
\begin{minipage}{0.8\textwidth}
    \centering
    \includegraphics[width=1.0\textwidth]{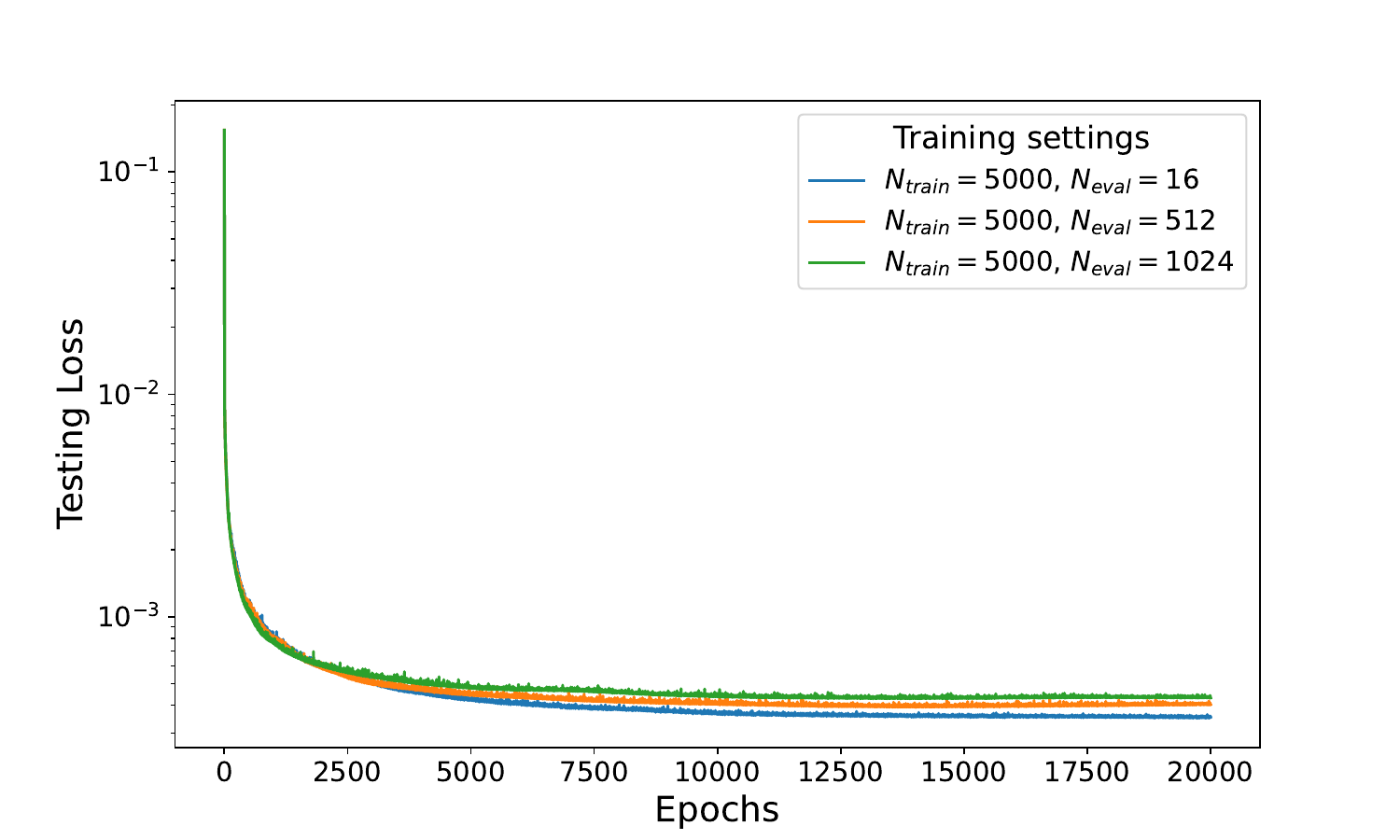}\\
    (b)
\end{minipage}
\caption{For the heat equation: (a) shows the training loss plot and (b) displays the test loss plot for a specific seed with $N_\text{train} = 5000$, varying the number of evaluation points used for the trunk network.}
\label{fig:Example3_Lossplots}
\end{figure}

\begin{figure}[H]
    \centering
    \includegraphics[width=\textwidth]{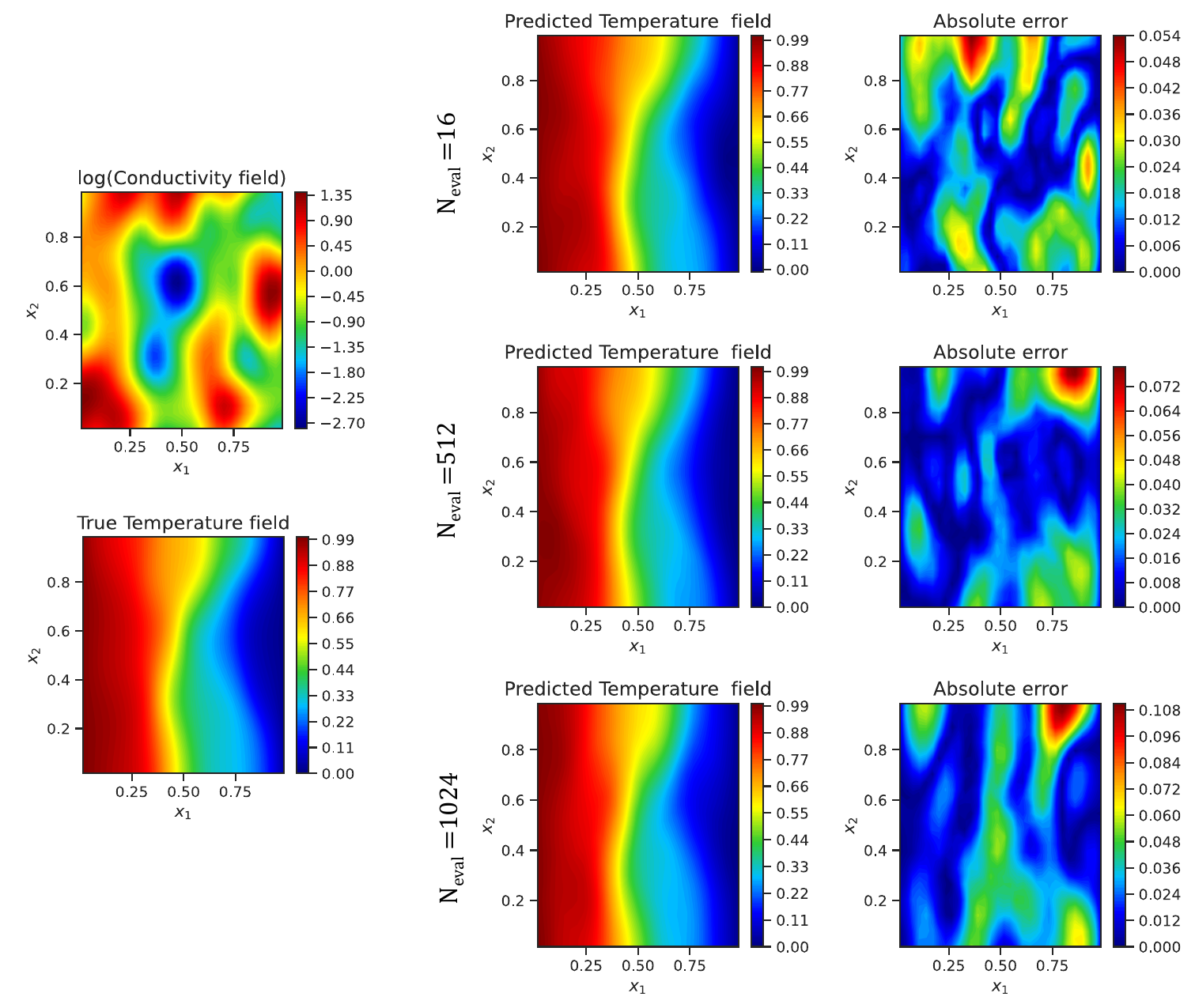}
    \caption{Heat equation: Comparison of all models for a representative test sample across different $N_\text{eval}$ values, with $N_\text{train} = 2500$.}
    \label{fig:Example3_sample_realization}
\end{figure}

As illustrated in Figure~\ref{fig:Example3_Boxplots}(a), the model performance with $N_\text{eval} =16$ and $512$ is comparable to or better than the performance with $N_\text{eval} = 1024$, which corresponds to evaluating the trunk network at all available evaluation points. This supports our hypothesis that randomizing the inputs to the trunk network effectively improves the training process, while significantly reducing the training time (see Figure~\ref{fig:Example3_Boxplots}(c)). 
Additionally, the loss plots in Figure~\ref{fig:Example3_Lossplots} show that training with a randomized trunk using $N_\text{eval} = 16$ or $512$ exhibits better generalization behavior than training with $N_\text{eval} = 1024$.
Figure~\ref{fig:Example3_sample_realization} illustrates a comparison of all models for a representative test sample with varying $N_\text{eval}$ values, where $N_\text{train} = 2500$.

% Ntrain=2500, test sample=380, seed=0

\section{Summary}\label{sec1.5}
\noindent

In this chapter, we have introduced a random sampling method designed to enhance the training efficiency of DeepONet and improve the generalization abilities of the framework. 
Our approach involves evaluating the trunk network at randomly selected domain points within the bounded domain during each iteration, rather than evaluating the loss at all available domain points. 
This method facilitates better exploration of the parameter space by the stochastic gradient descent (SGD) optimizer, potentially leading to a better local optimum. 
Additionally, this randomization approach significantly reduces training time due to smaller effective batch sizes in each iteration. 
We empirically validated the effectiveness of this method through three benchmark examples, performing ablation studies by varying the size of the labeled dataset and the number of evaluation points of the trunk network for each input function per iteration. 
Our results show that using a reasonable number of evaluation points, randomly selected, one can achieve comparable or even superior accuracy compared to evaluating the trunk network at every domain point.

The optimal number of evaluation points for minimizing generalization error is problem-dependent and remains a topic for future research. However, our findings suggest:
\begin{itemize}
    \item Using too few evaluation points can lead to poor generalization, as the DeepONet model does not see enough data in each iteration.
    \item Conversely, using too many evaluation points results in larger batch sizes, which can cause the model to get stuck in sharp minima, also leading to poor generalization.
\end{itemize}
Finding a balanced number of evaluation points is crucial for optimal performance. While our method shows promising results, several avenues for future research emerge, including developing methodologies to determine the optimal number of evaluation points for specific problem domains, and exploring adaptive sampling strategies that dynamically adjust the number of evaluation points during training.

\bibliographystyle{plain}
\bibliography{scibib}

\section*{Acknowledgements}
The authors would like to acknowledge computing support provided by the Advanced Research Computing at Hopkins (ARCH) core facility at Johns Hopkins University and the Rockfish cluster where all experiments were carried out. 

\section*{Funding}
S.K., L.G.B, and S.G. are supported by the U.S. Department of Energy, Office of Science, Office of Advanced Scientific Computing Research grant under Award Number DE-SC0024162.

\section*{Data and code availability}
All code and data accompanying this manuscript is publicly available at \url{https://github.com/Centrum-IntelliPhysics/DeepONet-Random-Sampling}.

\section*{Author contributions}
\noindent 
Conceptualization: S.K., L.G.B, S.G. \\
Investigation: S.K., L.G.B, S.G. \\
Visualization: S.K., L.G.B, S.G. \\
Supervision: L.G.B, S.G. \\
Writing—original draft: S.K. \\
Writing—review \& editing: S.K., L.G.B, S.G.

\end{document}